\documentclass[10pt,twocolumn,letterpaper]{article}

\usepackage{cvpr}              %

\usepackage[accsupp]{axessibility}  %

\usepackage{graphicx}
\usepackage{amsmath}
\usepackage{amssymb}
\usepackage{booktabs}

\usepackage[pagebackref,breaklinks,colorlinks]{hyperref}

\usepackage[capitalize]{cleveref}
\crefname{section}{Sec.}{Secs.}
\Crefname{section}{Section}{Sections}
\Crefname{table}{Table}{Tables}
\crefname{table}{Tab.}{Tabs.}

\def\cvprPaperID{7529} %
\def\confName{CVPR}
\def\confYear{2023}

\makeatletter
\apptocmd\@maketitle{{\teaserfigure{}\par}}{}{}
\makeatother

\def\myshift#1{\raisebox{0.5ex}}
\newcommand{\teasernobox}{
\begin{subfigure}[b]{\linewidth}
    \centering
	\includegraphics[width=\linewidth, ,trim={0 7.2cm 0 5.5cm},clip]{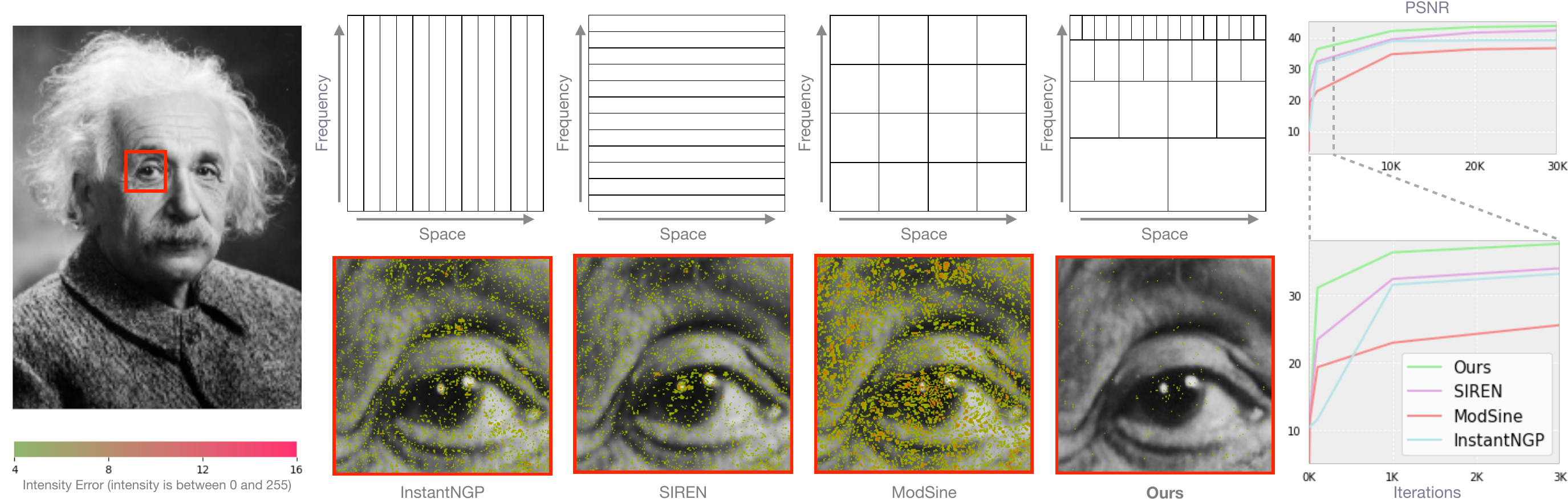} %
\end{subfigure}
}

\newcommand{\teaserfigure}{
\vspace{-5mm}
\captionsetup{type=figure}

\teasernobox

\vspace{-2mm}
\setcounter{figure}{0} %
\captionsetup{type=figure}
\captionof{figure}{
{\bf Teaser} --
We propose neural Fourier filter bank to perform spatial and frequency-wise decomposition jointly, inspired by wavelets.
Our method provides significantly improved reconstruction quality given the same computation and storage budget, as represented by the PSNR curve and the error image overlay.
Relying only on space partitioning without frequency resolution (InstantNGP)~\cite{muller2022instant}  or frequency encodings without space resolution (SIREN)~\cite{sitzmann2020implicit} provides suboptimal performance and convergence. 
Simply considering both (ModSine)~\cite{mehta2021modulated} enhances scalability when applied to larger scenes, but not in terms of quality and convergence.
}
\vspace{4mm}
\label{fig:teaser}
}
\vspace{-1em}

\usepackage[dvipsnames]{xcolor}
\usepackage{multirow}
\usepackage{mathabx}
\usepackage{enumitem}

\definecolor{turquoise}{cmyk}{0.65,0,0.1,0.1}
\definecolor{purple}{rgb}{0.65,0,0.65}
\definecolor{dark_green}{rgb}{0, 0.5, 0}
\definecolor{orange}{rgb}{0.8, 0.6, 0.2}
\definecolor{red}{rgb}{0.8, 0.2, 0.2}
\definecolor{blue}{rgb}{0, 0, 1}
\definecolor{brown}{rgb}{0.5, 0.16, 0.16}
\definecolor{black}{rgb}{0, 0, 0}

\newcommand{\Figure}[1]{\cref{fig:#1}}
\newcommand{\Table}[1]{\cref{tab:#1}}
\newcommand{\Section}[1]{\cref{sec:#1}}

\renewcommand{\paragraph}[1]{\vspace{0.3em}\noindent\textbf{#1}}

\newcommand{\supp}{\texttt{supplementary material\xspace}}

\newcommand{\bx}{\mathbf{x}}
\newcommand{\by}{\mathbf{y}}
\newcommand{\bw}{\mathbf{w}}
\newcommand{\gridfeat}{\mathbf{v}}
\newcommand{\real}{\mathbb{R}}

\newcommand{\linout}{\mathbf{f}}
\newcommand{\interout}{\mathbf{g}}
\newcommand{\levelout}{\mathbf{o}}

\newcommand{\weight}{\mathbf{W}}
\newcommand{\bias}{\mathbf{b}}
\newcommand{\scaling}{\alpha}

\newcommand{\singlelayer}{\boldsymbol{\kappa}}
\newcommand{\lookuptable}{\boldsymbol{\Phi}}
\newcommand{\interp}{\boldsymbol{\varphi}}
\newcommand{\posenc}{\boldsymbol{\gamma}}
\newcommand{\finalestimate}{\mathcal{F}}

\newcommand{\transmission}{\mathcal{T}}

\newcommand{\bc}{\mathbf{c}}
\newcommand{\Color}{C}
\newcommand{\ray}{\mathbf{r}}
\newcommand{\Rays}{\mathfrak{R}}

\newcommand{\layerseries}{\mathfrak{L}}
\newcommand{\layer}{L}
\newcommand{\outputseries}{\mathfrak{O}}
\newcommand{\outputlayer}{O}

\newcommand{\tablenumfeat}{T}
\newcommand{\tablefeatdim}{F}
\newcommand{\gridres}{N}
\newcommand{\primenum}{\Pi}

\begin{document}

\title{Neural Fourier Filter Bank}

\author{%
Zhijie Wu\qquad
Yuhe Jin\qquad
Kwang Moo Yi\\[.2mm]
University of British Columbia\\
{\tt\small \{zhijiewu, yuhejin,  kmyi\}@cs.ubc.ca}}

\maketitle

\begin{abstract}
We present a novel method to provide efficient and highly detailed reconstructions.
Inspired by wavelets, we learn a neural field that decompose the signal both spatially and frequency-wise.
We follow the recent grid-based paradigm for spatial decomposition, but unlike existing work, 
encourage specific frequencies to be stored in each grid via Fourier features encodings.
We then apply a multi-layer perceptron with sine activations, taking these Fourier encoded features in at appropriate layers so that higher-frequency components are accumulated on top of lower-frequency components sequentially, which we sum up to form the final output.
We demonstrate that our method outperforms the state of the art regarding model compactness and convergence speed on multiple tasks: 2D image fitting, 3D shape reconstruction, and neural radiance fields.
Our code is available at 
\url{https://github.com/ubc-vision/NFFB}.
\end{abstract}

\section{Introduction}
\label{sec:intro}

Neural fields~\cite{nfsurvey} have recently been shown to be highly effective for various tasks ranging from 2D image compression~\cite{zhang2022implicit,dupont2022coin++}, image translation~\cite{skorokhodov2021adversarial,chen2021learning}, 3D reconstruction~\cite{sitzmann2019scene,peng2020convolutional}, to neural rendering~\cite{mildenhall2020nerf,barron2021mip,muller2022instant}. 
Since the introduction of early methods~\cite{sitzmann2019scene,park2019deepsdf,mildenhall2020nerf}, efforts have been made to make neural fields more efficient and scalable.
Among various extensions, we are interested in two particular directions:
those that utilize spatial decomposition in the form of grids~\cite{muller2022instant,chen2022tensorf, takikawa2021neural} that allow fast training and level of detail;
and those that encode the inputs to neural fields with high-dimensional features via frequency transformation such as periodic sinusoidal representations~\cite{mildenhall2020nerf,tancik2020fourier,sitzmann2020implicit} that fight the inherent bias of neural fields that is towards low-frequency data~\cite{tancik2020fourier}.
The former drastically reduced the training time allowing various new application areas~\cite{tineuvox,tancik2022block,xiangli2021citynerf,yin20213dstylenet}, while the latter has now become a standard operation when applying neural fields.

While these two developments have become popular, a caveat in existing works is that they do not consider the two together---all grids are treated similarly and interpreted together by a neural network.
We argue that this is an important oversight that has a critical outcome.
For a model to be efficient and accurate, different grid resolutions should focus on different frequency components that are properly localized.
While existing grid methods that naturally localize signals—can learn to perform this frequency decomposition, relying purely on learning may lead to sub-optimal results as shown in \Figure{teaser}.
This is also true when locality is not considered, as shown by the SIREN~\cite{sitzmann2020implicit} example.
Explicit consideration of both together is hence important.

This caveat remains true even for methods that utilize both grids and frequency encodings for the input coordinates~\cite{muller2022instant} as grids and frequency are not linked, and it is up to the deep networks to find out the relationship between the two.
Thus, there has also been work that focuses on jointly considering both space and frequency~\cite{hertz2021sape,mehta2021modulated}, but these methods are not designed with multiple scales in mind thus single-scale and are designed to be non-scalable.
In other words, they can be thought of as being similar to short-time Fourier transform in signal processing.

\begin{figure}
    \centering
    \includegraphics[width=\linewidth,trim={2.6cm 7.3cm 4.7cm 5.6cm},clip]{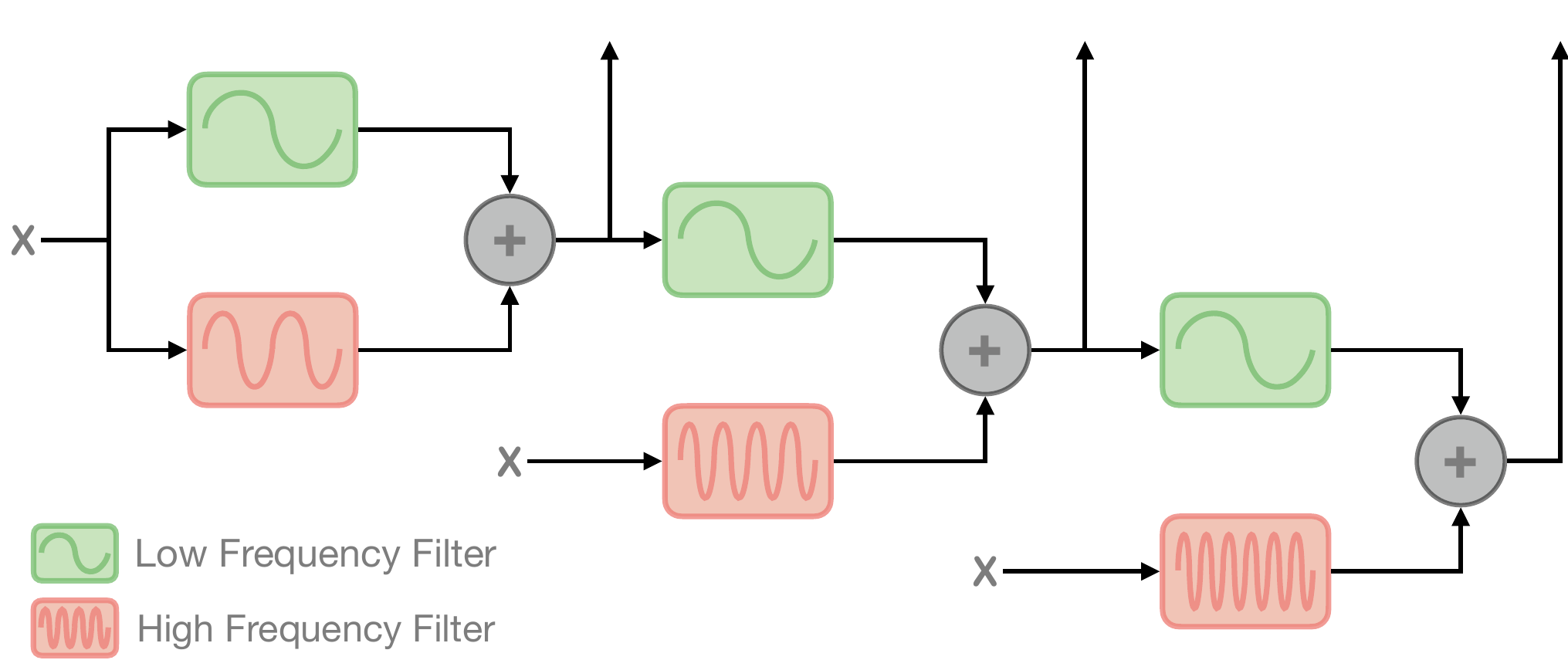}
    \caption{
    {\bf A wavelet-inspired framework} -- 
    In our framework, given a position $\bx$, low- and high-frequency filters are used to decompose the signal, which is then reconstructed by accumulating them and using the intermediate outputs as shown.
    Here, we utilize a multi-scale grid to act as if they store these high-frequency filtering outcomes at various spatially decomposed locations.
    }
    \label{fig:overview}
\end{figure}

Therefore, in this work, we propose a novel neural field framework that decomposes the target signal
in both space and frequency domains simultaneously, analogous to the traditional wavelet decomposition~\cite{shannon1949communication}; see \Figure{teaser}.
Specifically, a signal is decomposed jointly in space and frequency through low- and high-frequency filters as shown in \Figure{overview}.
Here, our core idea is to realize these filters conceptually as a neural network.
We implement the low-frequency path in the form of Multi-Layer Perceptrons~(MLP), leveraging their frequency bias~\cite{tancik2020fourier}.
For the high-frequency components, we implement them as lookup operations on grids, 
as the grid features can explicitly enforce locality over a small spatial area and facilitate learning of these components. 
This decomposition is much resemblant of filter banks in signal processing, thus we name our method neural Fourier filter bank.

In more detail, we utilize the multi-scale grid structure as in~\cite{muller2022instant,takikawa2021neural,hu2022efficientnerf}, but with a twist---we apply frequency encoding in the form of Fourier Features \emph{just before} the grid features are used.
By doing so, we convert the linear change in grid features that arise from bilinear/trilinear interpolation to appropriate frequencies that should be learned at each scale level.
We then compose these grid features together through an MLP with sine activation functions, which takes these features as input at each layer, forming a pipeline that sequentially accumulates higher-frequency information as composition is performed as shown in \Figure{overview}.
To facilitate training, we initialize each layer of the MLP with the target frequency band in mind.
Finally, we sum up all intermediate outputs together to form the estimated field value.

We demonstrate the effectiveness of our method under three different tasks: 2D image fitting, 3D shape reconstruction, and Neural Radiance Fields (NeRF).
We show that our method achieves a better trade-off between the model compactness versus reconstruction quality than the state of the arts.
We further perform an extensive ablation study to verify where the gains are coming from.

To summarize, our contributions are as follows:
\vspace{-.5em}
\begin{itemize}[leftmargin=*]
\setlength\itemsep{-.1em}
    \item we propose a novel framework that decomposes the modeled signal both spatially and frequency-wise;
    \item we show that our method achieves better trade-off between quality and memory on 2D image fitting, 3D shape reconstruction, and Neural Radiance Fields (NeRF);
    \item we provide an extensive ablation study shedding insight into the details of our method.
\end{itemize}

\section{Related Work}
\label{Sec:related_work}

Our work is in line with those that apply neural fields to model spatial-temporal signals~\cite{mescheder2019occupancy,park2019deepsdf,chen2019learning,mildenhall2020nerf,levoy1996light,qi2016volumetric,chen2022mobilenerf}. 
In this section, we survey representative approaches on neural field modeling~\cite{liu2018intriguing,bello2019attention,park2019deepsdf,tancik2020fourier,muller2022instant,takikawa2021neural} and provide an overview of work on incorporating the wavelet transform into deep network designs~\cite{de2016compression,isik2021lvac,gal2021swagan}.

\paragraph{Neural fields.}
A compressive survey can be found in \cite{nfsurvey}.
Here we briefly discuss representative work.
While existing methods have achieved impressive performance on modeling various signals that can be represented as fields~\cite{park2019deepsdf, peleg2019net, mescheder2019occupancy, genova2019learning, deng2020cvxnet, genova2020local, takikawa2021neural,muller2022instant}, neural fields can still fall short of representing the fine details ~\cite{genova2020local}, or incur high computational cost due to model complexity~\cite{jiang2020local}. 
Prior works attempt to solve these problems by \textit{frequency transformations}~\cite{sitzmann2020implicit,tancik2020fourier,mildenhall2020nerf} and \textit{grid-based encodings}~\cite{takikawa2021neural,genova2020local,muller2022instant}.

For \textit{frequency transformations}~\cite{muller2022instant}, Vaswani \etal~\cite{vaswani2017attention} encode the input feature vectors into a high-dimension latent space through a sequence of periodic functions. 
Tancik \etal~\cite{tancik2020fourier} carefully and randomly choose the frequency of the periodic functions and reveal how they affect the fidelity of results.
Sitzmann \etal~\cite{sitzmann2020implicit} propose to use periodic activation functions instead of encoding feature vectors.
~\cite{lindell2021bacon,fathony2021multiplicative} further push analysis in terms of the spectral domain with a multi-scale strategy, 
improving the capability in modeling band limited signals in one single model.
To further understand the success of these methods,~\cite{yuce2022structured,benbarka2022seeing} analyze the implicit representations from the perspective of a structured dictionary and Fourier series, respectively.

For \textit{grid-based encodings}~\cite{muller2022instant,takikawa2021neural}, the core idea is to encode the input to the neural field
by interpolating a learnable basis consisting of grid-point features (space partitioning).
A distinctive benefit of doing so is that one can trade memory for faster training---bigger networks can be used to represent complex scenes, as long as the entire grid used is within memory.
To reduce this memory footprint, compact hash tables~\cite{muller2022instant} and volumetric matrix decomposition~\cite{chen2022tensorf} have been introduced.
These recent methods, however, do not, at the very least explicitly, consider how grid resolutions and frequency interact.

Thus, 
some works try to combine both directions.
For example, SAPE~\cite{hertz2021sape} progressively encodes the input coordinates by attending to time-spatial information jointly.
Mehta \etal~\cite{mehta2021modulated} decompose the inputs into patches, which are used to modulate the activation functions. 
They, however, utilize a \emph{single} space resolution, limiting their modeling capability. 
Instead, we show that by using multiple scale levels, and a framework that takes into account the frequencies that are to be associated with these levels, one can achieve faster convergence with higher accuracy.

\paragraph{Wavelets in deep nets.}
The use of wavelet transforms has been well-studied in the deep learning literature.
For example, they have been used for wavelet-based feature pooling operations~\cite{gao2016hybrid,liu2019multi,williams2018wavelet}, for the improvements on style transfer~\cite{yoo2019photorealistic,gal2021swagan}, for denoising~\cite{liu2020wavelet}, for medical analysis~\cite{kang2017deep}, and for image generation~\cite{huang2019wavelet,wang2020multi,phung2022wavelet,liu2019attribute,wang2020multi}.
Recently, Liang \etal~\cite{liang2021reproducing} reproduce wavelets through linearly combining activation functions. 
Gauthier \etal~\cite{gauthier2022parametric} introduce wavelet scattering transform  to create geometric invariants and deformation stability.
Phung \etal~\cite{phung2022wavelet} use Haar wavelets with diffusion models to accelerate convergence.
In the 3D vision domain, De Queiroz \etal~\cite{de2016compression} propose a transformation that
resembles an adaptive variation of Haar wavelets to facilitate 3D point cloud compression.
Isik \etal~\cite{isik2021lvac} directly learn trainable coefficients of the hierarchical Haar wavelet transform, reporting impressive compression results. 
Concurrently, Rho\etal~\cite{rho2023masked} propose using wavelet coefficients to improve model compactness.
While our work shares a similar spirit as those that utilize wavelets, to the best of our knowledge, ours is the first work aimed at a general-purpose neural field architecture that jointly and explicitly models the spatial and frequency domains.

\begin{figure*}
    \centering
    \includegraphics[width=\linewidth,trim={0.3cm 10.3cm 5cm 2.9cm},clip]{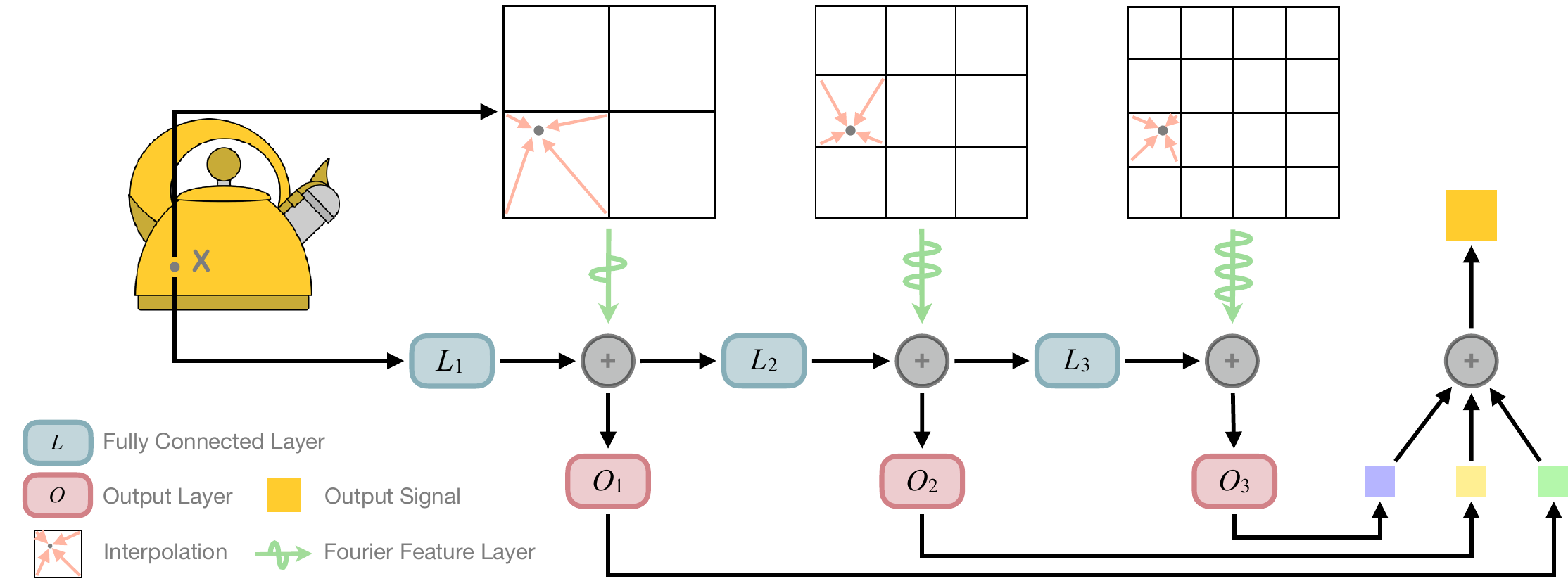}
    \vspace{-1.0em}
    \caption{
    {\bf Framework overview} --
    Based on the input query, \eg the position $x$, our neural Fourier filter bank uses both a grid and a Multi-Layer Perceptron (MLP) to \emph{compose} the final estimate.
    Specifically, grid features are extracted via interpolation at multiple scale levels, which are then encoded to appropriate frequencies for each layer via the Fourier Feature layers.
    The MLP uses these encoded features as the higher-frequency component in \Figure{overview}, while the earlier layer outputs as the lower frequency ones, similarly to wavelet filter banks.
    Intermediate outputs are then aggregated as the final estimate.
    }
    \vspace{-0.8em}
    \label{fig:detailedframework}
\end{figure*}

\section{Method}
\label{sec:method}

In this work, we aim for a multi-resolution grid-based framework that also ties in the frequency space to these grids, as is done with wavelets, and an architecture to effectively reconstruct the original signal.
As shown in \Figure{overview}, we construct our pipeline, neural Fourier filter bank, composed of two parts:
a Fourier-space analogous version of grid features 
(\Section{grid_feats});
and an MLP that composes the final signals from these grid values
(\Section{mlp_net}). 
We discuss these in more detail in the following subsections.

\subsection{The Fourier grid features} 
\label{sec:grid_feats}

As discussed earlier in \Section{intro}, 
we use a grid setup to facilitate the learning of high-frequency components via locality.
Specifically,
we aim for each grid level in the multi-grid setup to store different frequency bands of the field that we wish to store in the neural network.
The core idea in how we achieve this is to combine the typical grid setup used by, \eg~\cite{muller2022instant}, with Fourier features~\cite{tancik2020fourier}, which we then initialize appropriately to naturally encourage a given grid to focus on certain frequencies.
This is analogous to how one can control the frequency details of a neural field by controlling the Fourier feature~\cite{tancik2020fourier} encoding of the input coordinates, but here we are applying it to the grid features.

In more detail, the grid feature at the $i$-th level is defined as a continuous mapping from the input coordinate $\bx \in \real^n$ to $m$ dimension feature space:
\begin{equation}
\label{eq:grid_mapping}
\singlelayer_i: \real^n \to \real^m
.
\end{equation}
We set $n=$ 2, 3 for 2D images and 3D shapes respectively.
As shown in \Figure{detailedframework}, $\singlelayer_i$ consists of two parts: 
a lookup table $\lookuptable_i$ which has $\tablenumfeat_i$ feature vectors with dimensionality $\tablefeatdim$;
and a Fourier feature layer~\cite{tancik2020fourier} $\Omega_i$.

\paragraph{Multi-scale grid.}
We apply a trainable hash table~\cite{muller2022instant} to implement $\lookuptable_i$ for a better balance between performance and quality.
For the $i$-th level, we store the feature vectors at the vertices of a grid, the resolution of which $\gridres_i$ is chosen manually. 
To utilize this grid in a continuous coordinate setup, one typically performs linear interpolation~\cite{takikawa2021neural,muller2022instant}.
Hence, for a continuous coordinate $\bx$, to get the grid points,
for each dimension
we first scale $\bx$ by $\gridres_i$ before rounding down and up, which we write with a slight abuse of notation (ignoring dimensions) as:
\begin{equation}
\left \lfloor \bx_i \right \rfloor = \left \lfloor \bx \cdot \gridres_i \right \rfloor, \left \lceil \bx_i \right \rceil  = \left \lceil \bx \cdot \gridres_i \right \rceil
.
\end{equation}
Here, $\left \lfloor \bx_i \right \rfloor$ and $\left \lceil \bx_i \right \rceil$, for example occupies a voxel with $2^n$ integer vertices.
As in \cite{muller2022instant}, we then map each corner vertex to an entry in the matching lookup table, using a spatial hash function~\cite{muller2022instant,teschner2003optimized} as:
\begin{equation}
h(\bar{\bx}) = \left \{ \bigwedge_{i=1}^{n} \bar{\bx}_i \cdot \primenum_i \right \} \enspace \mathrm{mod \enspace  T_i}
,
\end{equation}
where $\bar{\bx}$ represents the position of a specific corner vertex, $\bigwedge$ denotes the bit-wise XOR operation and $\primenum_i$ are unique, large prime numbers.
As in~\cite{muller2022instant}, we choose $\primenum_1 = 1$, $\primenum_2 = 2 654 435 761$ and $\primenum_3 = 805 459 861$. 

Finally, for $\bx$, we perform linear interpolation for its $2^n$ corner feature vectors based on their relative position to $\bx$ within its hypercube as $\bw_i=\bx_i - \left \lfloor \bx_i \right \rfloor$. 
Specifically, we use bilinear interpolation for 2D image fitting and trilinear interpolation for 3D shape modeling. 
We denote the output features through the linear interpolation over the lookup table $\lookuptable_i$ as $\interp (\bx; \lookuptable_i)$.

It is important to note that this linear interpolation operation makes these features behave similarly to how the input coordinates affect the neural field output~\cite{tancik2020fourier}---introducing bias toward slowly changing components.
Thus, in order for each grid level to focus on appropriate frequency bands it is necessary to explicitly take this into account.

\paragraph{Converting grid features to Fourier features.}
Then, to associate the spatial area with the specific frequency level, we apply Fourier feature encoding to $\gridfeat_i=\interp (\bx; \lookuptable_i)$ before we utilize them:
\begin{equation}
\begin{aligned}
\posenc_i (\gridfeat_i) = [\sin (2\pi \cdot B_{i,1} \cdot \gridfeat_i^\top), \dots , \sin (2\pi \cdot B_{i,m} \cdot \gridfeat_i^\top) ]^\top
,
\end{aligned}
\end{equation}
where $\left \{ B_{i,1}, B_{i,2}, \cdots, B_{i,m} \right \}$ means trainable frequency transform coefficients on $i$-th level.
We then utilize $\posenc_i (\gridfeat_i)$ in our network that converts these into desired field values.

Importantly, we directly associate the frequency band on the $i$-th level with desired grid size
by explicitly initializing $\left \{ B_{i,1}, B_{i,2}, \cdots, B_{i,m} \right \}$ with adaptive Gaussian distribution variance similarly to Gaussian mapping~\cite[Sec.~6.1]{tancik2020fourier}.
We choose to initialize with different variances, as it is difficult to set a specific frequency range for a given grid \emph{a priori}.
Instead of trying to set a proper range that is hard to accomplish, we initialize finer grids with larger variance and naturally bias finer grids towards higher frequency components since the multiplier for $\gridfeat$ will then be larger---they will be biased to converge to larger frequencies~\cite{hertz2021sape}.

\subsection{Composing the field value} 
\label{sec:mlp_net}

To compose the field values from our Fourier grid features, we 
start from two important observations:
\vspace{-.5em}
\begin{itemize}[leftmargin=*]
\setlength\itemsep{-.2em}
    \item The stored Fourier grid features at different layers, after going through a deep network layer for interpretation, are not orthogonal to each other. 
    This calls for the need for learned layers 
    when aggregating features from different levels so that this non-orthogonality is mitigated.
    \item The Fourier grid features should be at a similar `depth' so that they are updated simultaneously.
    This makes residual setups preferable.
\end{itemize}
We thus utilize an MLP, which takes in the Fourier grid features at various layers.
As shown in \Figure{detailedframework}, each layer takes in features from the previous layer, as well as the Fourier grid features, then either passes it to the next layer or to an output feature that is then summed up to form a final output.

Mathematically, denoting the MLP as a series of fully-connected layers $\layerseries=\left \{ \layer_1, \layer_2,\cdots \right \}$, we write
\begin{equation}
\linout_i = \sin (\scaling_i\cdot \weight_i \interout_{i-1} + \bias_i), \quad \interout_i = \linout_i + \posenc_i (\gridfeat_i),
\end{equation}
where $\weight_i$ and $\bias_i$ are trainable weight and bias in the $i$-th layer $\layer_i$, and $\scaling_i$ is the scaling factor for this layer that control the frequency range that this layer focuses on, which is equivalent to the $w_0$ hyperparamter in SIREN~\cite{sitzmann2020implicit}.
Note here that $\linout_i$ corresponds to the output of the lower-frequency component, and the Fourier grid features $\posenc_i(\gridfeat_i)$ are the higher-frequency ones in \Figure{overview}.
For the first layer, as there is no earlier level, we use the input position $x$. 
Thus,
\begin{equation}
\linout_1 = \sin (\scaling_1\cdot\weight_1 \bx + \bias_1), \quad \interout_1 = \linout_1 + \posenc_1 (\gridfeat_1)
.
\end{equation}
Then, with $\interout_i$, we construct the per-level outputs $\levelout_i = \weight_i^o \interout_i + \bias_i^o$ with 
output layers
$\outputseries=\left \{ \outputlayer_1, \outputlayer_2,\cdots \right \}$ with
another trainable parameters set $\{\weight_1^o, \weight_2^o, \cdots, \bias_1^o, \bias_2^o, \cdots\}$.
We then sum up $\levelout_i$ to obtain the final estimated field value as 
$\finalestimate(\bx)=\sum_{i=1} \levelout_i$.

\paragraph{Importance of the composition architecture.}
A simpler alternative to composing the field signal estimate would be to simply use Fourier grid features in an existing pipeline~\cite{takikawa2021neural,muller2022instant} that utilizes grids.
However, as we will show in \Section{ablation}, this results in consistently inferior performance compared to our method of composition.

\subsection{Implementation details} 
\label{sec:implementation}

Depending on the target applications, some implementation details vary---the loss function, the number of training iterations, and the network capacity are task dependant and we elaborate on them later in their respective subsections.
Other than the task-specific components we keep the same training setup for all experiments.
We implement our method in PyTorch~\cite{pytorch}.
We use the Adam optimizer~\cite{kingma2014adam} with default parameters $\beta_1 = 0.9$ and $\beta_2 = 0.99$.
We use a learning rate of $10^{-4}$, and decay the learning rate to half every 5,000 iterations.
We set the dimension of grid features as $F=2$.
We train our method on a single NVidia RTX 3090 GPU.
Here, for brevity, we note only the critical setup for each experiment.
For more details on the architectures and the hyperparameter settings, please see the \supp.

\section{Experimental Results}
\label{Sec:experiments}

We evaluate our method on three different tasks: 2D image fitting (\Section{results2d}), 3D shape reconstruction using signed distance functions (\Section{results3d}), and novel view synthesis using NeRF (\Section{resultsnerf}). Ablation study is shown in \Section{ablation}.
More experiment discussions can be found in the appendix.

\subsection{2D Image Fitting}
\label{sec:results2d}

\begin{figure*}
    \centering
    \includegraphics[width=\linewidth,trim={0 3cm 10.6cm 3.4cm},clip]{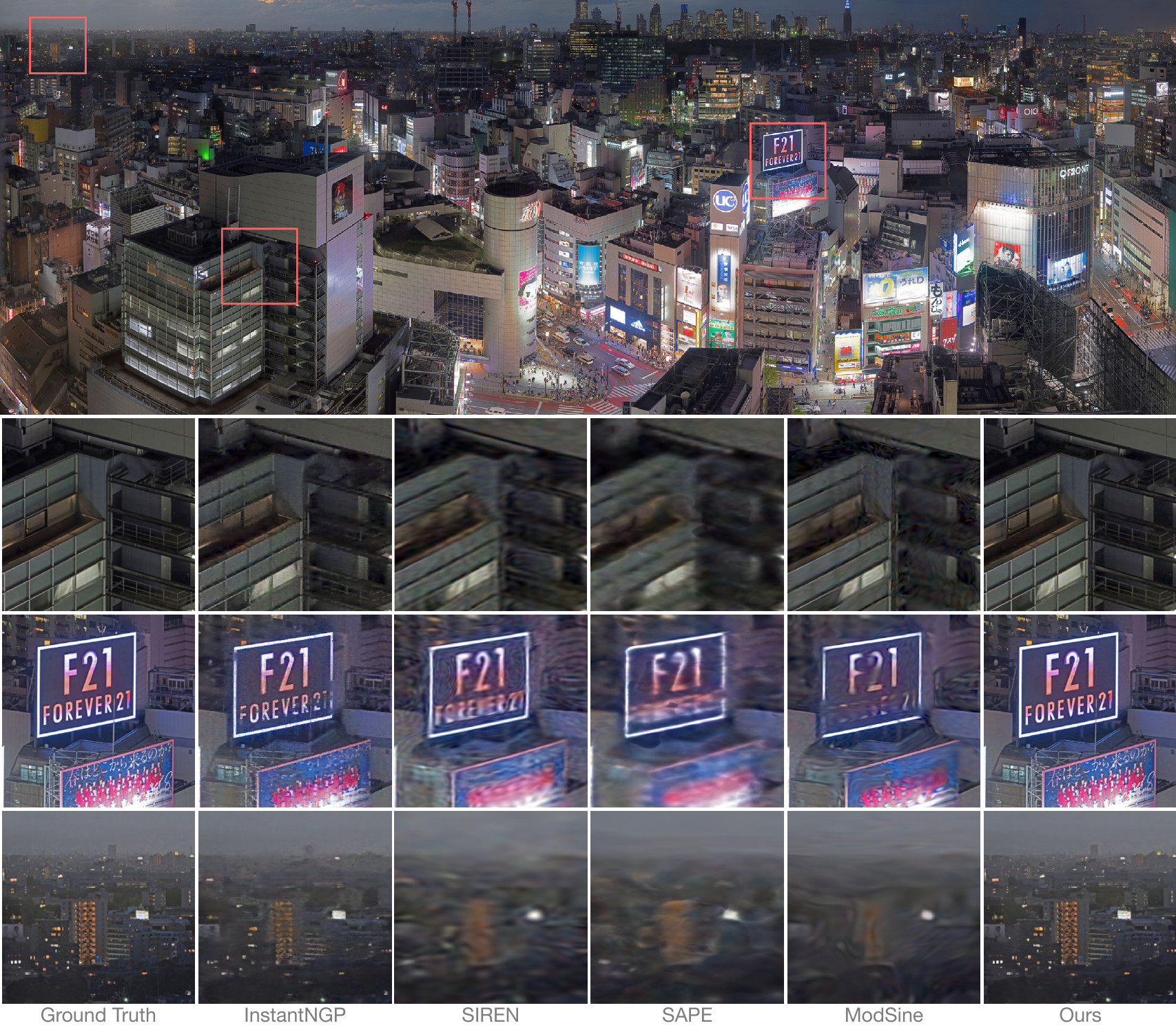}
    \vspace{-1em}
    \caption{
    {\bf 2D Fitting} -- 
    Qualitative results for the Tokyo image.
    Our method provides the best reconstruction quality at various scale levels, from nearby regions to far away ones, demonstrating the importance of considering both space and frequency jointly.
    }
    \label{fig:2D_fitting}
\end{figure*}
\begin{table}

\begin{center}

\resizebox{\linewidth}{!}{

\begin{tabular}{@{}l c c c c c c c c c@{}}
    \toprule
     & \multicolumn{4}{c}{Tokyo} & \multicolumn{4}{c}{Albert}\\
    \cmidrule(l){2-5} \cmidrule(l){6-9} 
    & Size~(MB)$\downarrow$ & PSNR$\uparrow$ & SSIM$\uparrow$ & LPIPS$\downarrow$ & Size~(MB)$\downarrow$ & PSNR$\uparrow$   & SSIM$\uparrow$  & LPIPS$\downarrow$ \\
    \midrule
    InstantNGP~\cite{muller2022instant}  & 36.0 & 33.38 & 0.9452 & 0.201 & 3.7 & 41.61 & 0.9623 & 0.152\\
    SIREN~\cite{sitzmann2020implicit} & 5.2 & 28.52 & 0.8921 & 0.474 & 5.0 & 42.51 & 0.9661 & 0.478 \\
    SAPE~\cite{hertz2021sape} & \textbf{3.2} & 21.64 & 0.5357 & 0.745 & \textbf{3.2} & 34.26 & 0.9219 & 0.399 \\
    ModSine~\cite{mehta2021modulated} & 3.5 & 23.23 & 0.7587 & 0.607 & 4.2 & 36.74 & 0.9184 & 0.438 \\
    \midrule
    Ours & 10.0 & \textbf{33.62} & \textbf{0.9555} & \textbf{0.141} &  3.7 & \textbf{43.83} & \textbf{0.9763} & \textbf{0.142} \\
    \bottomrule
\end{tabular}
}
\end{center}
\vspace{-1em}
\caption{
\textbf{2D Fitting} --
We report the reconstruction comparisons
in terms of Peak Signal-to-Noise Ratio (PSNR), Structural Similarity Index Metric (SSIM)~\cite{hore2010image} and Learned Perceptual Image Patch Similarity (LPIPS)~\cite{zhang2018unreasonable}.
Our method provides the best trade-off between model size and reconstruction quality.
\label{tab:2D_fitting}
}
\end{table}

We first validate the effectiveness of our method in representing large-scale 2D images.
For all models, we train them with the mean squared error. 
Hence, our loss function for this task is
\begin{equation}
	\mathcal{L}_{img} = \left \| \by - \by_{{\scriptsize gt}} \right \|^2_2
 ,
\end{equation}
where $\by$ is the neural field estimate and $\by_{{\scriptsize gt}}$ is the ground-truth pixel color.

\paragraph{Data.}
To keep our experiments compatible with existing work, we follow ACORN~\cite{martel2021acorn} and evaluate each method on two very high-resolution images.
The first image is a photo of `Einstein'\footnote{Collected from \href{https://github.com/NVlabs/tiny-cuda-nn}{https://github.com/NVlabs/tiny-cuda-nn.}}, already shown in \Figure{teaser}.
This image has a resolution of $3250 \times 4333$ pixels, with varying amounts of details in different regions of the image, making it an interesting image to test how each model is capable of representing various levels of detail---background is blurry and smooth, while the eye and the clothes exhibit high-frequency details. 
Another image is a photo of the nightscape of `Tokyo'~\cite{martel2021acorn}
with a resolution of $6144 \times 2324$, where near and far objects provide a large amount of detail at various frequencies.

\paragraph{Baselines.}
We compare our method against four different baselines designed for this task:
InstantNGP~\cite{muller2022instant}, which utilizes grid based space partitions for the input;
SIREN~\cite{sitzmann2020implicit}, which resembles modeling the Fourier space;
and two methods (SAPE~\cite{hertz2021sape} and ModSine~\cite{mehta2021modulated}) that consider both the frequency and the space decomposition but not as in our method.
For all methods, we use the official implementation by the authors but change their model capacity (number of parameters, and grid/hash table size) and task-specific parameters.
Specifically, for SIREN, we set the frequency parameter $\omega_0 = 30.0$ and initialize the network with 5 hidden layers with size $512 \times 512$. 
For SAPE, we preserve their original network size. 
For InstantNGP, we adjust its maximum hashtable size as $T=2^{17}$ and the grid level to $L=8$ for the `Einstein' image and set $T=2^{19}$ and $L=16$ for `Tokyo' to better cater to complex details.
To allow all models to fully converge, we report results after 50,000 iterations of training.

\paragraph{Results.}
We provide qualitative results for the `Tokyo' image in \Figure{2D_fitting}, and report the quantitative metrics in \Table{2D_fitting}.
As shown, our method provides the best tradeoff between model size and reconstruction quality, both in terms of Peak Signal-to-Noise Ratio (PSNR),  Structural Similarity Index Metric (SSIM)~\cite{hore2010image}, and Learned Perceptual Image Patch Similarity (LPIPS)~\cite{zhang2018unreasonable}.
Among these, note that the gap in performance is larger with SSIM and LPIPS, which better represents the local structure differences.
This is also visible in \Figure{2D_fitting}, where our method provides results that are nearly indistinguishable from the ground truth.

We note that the importance of considering both frequency and space is well exemplified in \Figure{2D_fitting}.
As shown, while InstantNGP provides good details for nearby regions (second row), as further away regions are investigated (third and last row), artifacts are more visible.
This demonstrates that even when multiscale grid is used, without consideration of the frequencies associated with these scales, results degrade.
Other baselines, SIREN, ModSine, and SAPE, are all single-scale and show results as if they are focusing on a single frequency band.
Ours on the other hand does not suffer from these artifacts.

\subsection{3D Shape Reconstruction}
\label{sec:results3d}

\begin{figure*}
    \centering
    \includegraphics[width=\linewidth,trim={0 12.9cm 4cm 2.7cm},clip]{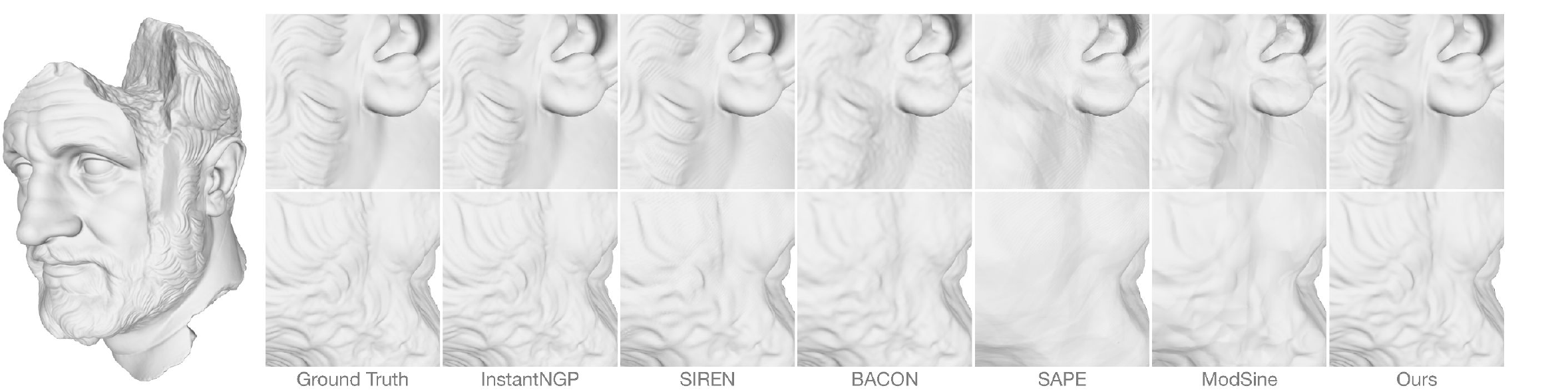}
    \vspace{-1em}
    \caption{
    {\bf 3D Fitting} -- Qualitative comparisons for the `Bearded Man' shape. 
    Our method is the most compact among the compared methods, and is capable of reconstructing both coarse and fine details without obvious artifacts.
    }
    \label{fig:3D_fitting}
\end{figure*}
\begin{table}

\begin{center}

\resizebox{\linewidth}{!}{

\begin{tabular}{@{}l c c c c c c c c @{}}
    \toprule
    & \multirow{2}{*}{Size~(MB)$\downarrow$} & \multicolumn{3}{c}{Asian Dragon} & \multicolumn{3}{c}{Bearded Man}\\
    \cmidrule(l){3-5} \cmidrule(l){6-8} 
    &  & F-score$\uparrow$ & IoU$\uparrow$  & Cham dist$\downarrow$ & F-score$\uparrow$ & IoU$\uparrow$  & Cham dist$\downarrow$\\
    \midrule
    InstantNGP \cite{muller2022instant}      & 46.5 & 0.8714 & \textbf{1.0}   & 0.00191 & \textbf{0.999} & 0.9970 & \textbf{0.00272}\\
    SIREN \cite{sitzmann2020implicit}    & 2.0  & 0.8593 & 0.998 & 0.00234 & 0.997 & 0.9951 & 0.00302\\
    BACON \cite{lindell2021bacon}    & 2.0  & \textbf{0.9200} & 0.995 & 0.00242 & 0.716 & 0.9932 & 0.00285\\
    SAPE \cite{hertz2021sape}     & 3.2     & 0.3210 & 0.959 & 0.00584 & 0.284 & 0.9837 & 0.00438\\
    ModSine \cite{mehta2021modulated} & 12.0 & 0.6892 & 0.995 & 0.00238 & 0.873 & 0.9952 & 0.00307\\
    \midrule
    Ours     & \textbf{1.4}  & 0.8717 & \textbf{1.0}   & \textbf{0.00189}  & \textbf{0.999} & \textbf{0.9985} & \textbf{0.00272}\\
    \bottomrule
\end{tabular}
}
\end{center}
\vspace{-1em}
\caption{
\textbf{3D Fitting} --
We report the Intersection over Union (IoU), F-Score and Chamfer distance (CD) after performing marching cubes to extract surfaces. 
Our method performs best, with the exception of F-score on `Asian Dragon', which is due to BACON preferring blobby output, as demonstrated by the higher Chamfer distance and worse IoU.
\label{tab:3D_fitting}
}

\end{table}

We further evaluate our method
on the task of representing 3D shapes as signed distance fields (SDF).
For this task, we use the square of the Mean Absolute Percentage Error (MAPE)~\cite{muller2022instant} as training objective, to facilitate detail modeling.
We thus train models by minimizing the loss:
\begin{equation}
\mathcal{L}_{sdf} = \left \| \by - \by_{gt} \right \|^2_2 / \left ( \epsilon + \left \| \by_{gt} \right \|^2_2 \right ),
\end{equation}
where $\epsilon$ denotes a small constant to avoid numerical problems, $y$ is the neural field estimate, and $y_{gt}$ is the ground-truth SDF value.

\paragraph{Data.}
For this task, we choose two standard textured 3D shapes for evaluation: 
`Bearded Man' (with 691K vertices and 1.38M faces); 
and `Asian Dragon' (3.6M vertices and 7.2M faces).
Both shapes exhibit coarse and fine geometric details.
When training with these shapes, we sample 3D points $x \in R^3$ with a 20/30/50 split---$20\%$ of the points are sampled uniformly within the volume, $30\%$ of the points are sampled near the shape surface, and the rest sampled directly on the surface.

\paragraph{Baselines.}
We compare against the same baselines as in \Section{results2d}, and additionally BACON, which also utilizes frequency decomposition for efficient neural field modeling.
For BACON and SIREN, we use networks with 8 hidden layers and 256 hidden features, and again $\omega_0 = 30.0$ for SIREN. 
For ModSine, we set the grid resolution as $64 \times 64 \times 64$ and apply 8 hidden layers and 256 hidden features for both the modulation network and the synthesis network. 
For SAPE and InstantNGP, use the author-tuned defaults for this task.
All models are trained for 100K iterations for full training.

\paragraph{Results.}
We present our qualitative results in \Figure{3D_fitting} and report quantiative scores in \Table{3D_fitting}. 
To extract detailed surfaces from each implicit representation we apply marching cubes with a resolution of $1024^3$.
As shown, our method provides the best performance, while having the smallest model size.
Note that in \Table{3D_fitting} our results are worse in terms for F-score for the Asian Dragon, while the other metrics report performance comparable to InstantNGP with 30$\times$ smaller model size.
The lower F-score but higher Chamfer distance is due to our model having lower recall than BACON, which provides more blobby results, as demonstrate by the IoU and Chamfer distance metrics.
We also note that for the `Bearded Man', our method outperforms all other methods.

This difference in quantitative metrics is also visible in \Figure{3D_fitting}.
As shown, our method provides high-quality reconstruction for both zoomed-in regions, whereas other compared methods show lower-quality reconstructions for at least one of them.
For example, SIREN provides good reconstruction for the beard region (second row), but not for the region around the ears (top row), where sinusoidal artifacts are visible.
InstantNGP also delivers high-quality reconstruction for the `Bearded Man', but with much higher memory requirement.

\subsection{Novel View Synthesis}
\label{sec:resultsnerf}

\begin{figure*}
    \centering
    \includegraphics[width=\linewidth,trim={0.1 13cm 5.8cm 2.9cm},clip]{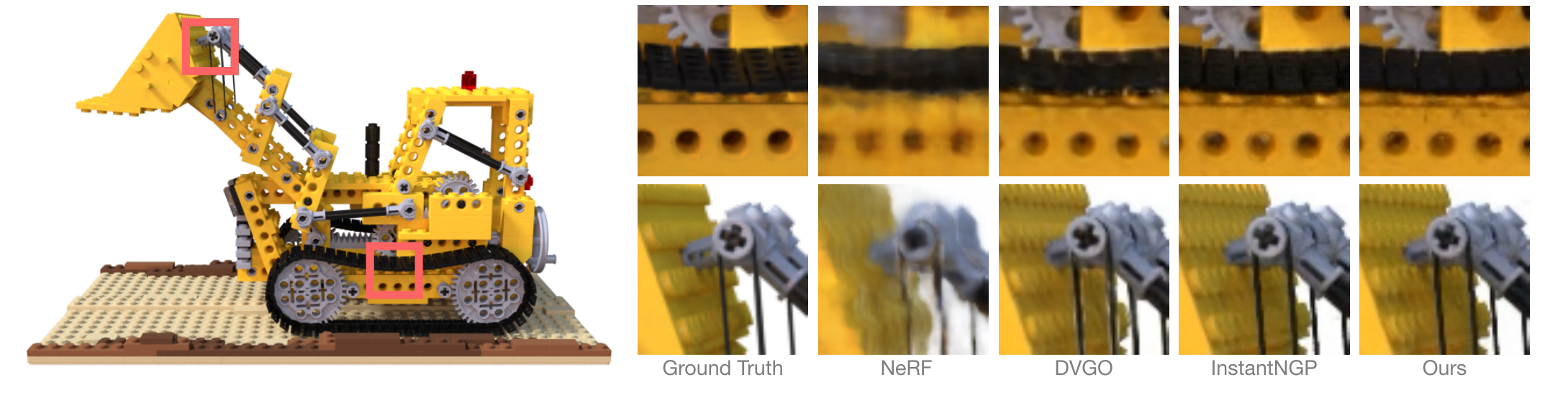}
    \vspace{-2.5em}
    \caption{
    {\bf Novel View Synthesis} -- 
    Although more compact, our method can synthesize comparable or better results.
    }
    \label{fig:NVS}
\end{figure*}
\begin{table}

\begin{center}
\resizebox{\linewidth}{!}{

\begin{tabular}{ c  c c c c c}
    \toprule
     & Steps  & Size~(MB) $\downarrow$ & Time $\downarrow$ & PSNR $\uparrow$ & SSIM $\uparrow$ \\
     \midrule
    NeRF\cite{mildenhall2020nerf} & 300k  & 5.0 & $>30$h & 31.01 & 0.947 \\
    \midrule
    Plenoxels \cite{yu2021plenoxels}& 128k & 778.1 & 11.4m & 31.71 & 0.958 \\
    DVGO \cite{SunSC22} & 30k &  612.1 & 15m & 31.95 & 0.957 \\
    InstantNGP \cite{muller2022instant} & 30k & 46.6 & 3.4m & 32.08 & 0.955 \\
    \midrule
    Ours & 30k  & 14.7 & 13.1m & 32.04 & 0.955 \\
    \bottomrule
\end{tabular}
}
\end{center}
\vspace{-1em}
\caption{
    {\bf Neural Radiance Fields (NeRF)} --
    We report the novel view rendering performance in terms of Peak Signal-to-Noise Ratio (PSNR) and Structural Similarity Index Metric (SSIM).
    Our method provides comparable rendering quality as the state of the art, while having the smallest size among the grid-based methods (middle rows) that provide fast training, providing the best trade-off between quality and model size.
    See the appendix for runtime discussions.
\label{tab:NVS}
}

\end{table}
\begin{figure*}
    \centering
    \vspace{-1em}
    \includegraphics[width=\linewidth,trim={0 10.2cm 0.3cm 6.2cm},clip]{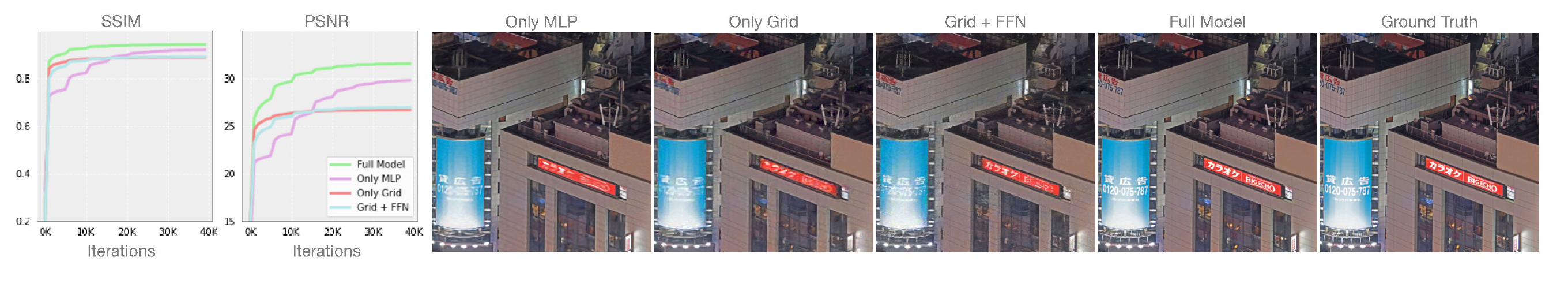}
    \vspace{-2.5em}
    \caption{
    {\bf Ablation study} --
    We compare against variants of our method with the Fourier grid feature and/or the proposed MLP composition architecture disabled.
    Having both components \emph{together} is critical for performance.
    }
    \label{fig:ablation_component}
\end{figure*}

As our last task, we apply our method to modeling Neural Radiance Fields (NeRF)~\cite{mildenhall2020nerf}.
Because we are interested in comparing the neural field architectures, not the NeRF method itself, we focus on the simple setup using the synthetic Blender dataset.

We train all architectures with a pure NeRF setup~\cite{mildenhall2020nerf}, where volumetric rendering is used to obtain pixel colors, which are then compared to ground-truth values for training.
Specifically, a pixel color is predicted as
\begin{equation}
\begin{aligned}
\hat{\Color} \left ( r \right ) &= \sum_{i=1}^{n} \transmission_i \left ( 1 - \mathrm{exp} (-\sigma_i \delta _i) \right ) \bc_i, \\
\transmission_i &= \mathrm{exp}(-\sum_{j=1}^{i-1} \sigma_j \delta_j),
\end{aligned}
\end{equation}
where $\bc_i$ and $\sigma_i$ denote the color and density estimated at the $i$-th queried location along the ray $\ray$ and $\delta_i$ is the distance between adjacent samples along a given ray.
Then, the mean-squared loss for training is:
\begin{equation}
\mathcal{L}_{rec} = \sum_{\ray \in \Rays } \left \| \hat{\Color}(r) - \Color_{gt}(r) \right \|^2_2
,
\end{equation}
where $\Rays$ is the whole ray set and $\Color_{gt}$ is the ground truth.

\paragraph{Adaptation.} 
For this task, we found that the complexity of the task, estimating both the color and the density, requires appending our pipeline with an additional MLP that decodes deep features into either the color or the density.
Thus, instead of directly outputting these values from our framework, we output a deep feature, which is then converted into color and density.
Specifically, as in NeRF~\cite{mildenhall2020nerf}, we apply two $64 \times 64$ linear layers to predict density value and a low-dimension deep feature, which is further fed into three $64 \times 64$ linear layers for RGB estimation.

\paragraph{Baselines.} 
We compare against five baselines:
NeRF~\cite{mildenhall2020nerf} which is utilizes the frequency domain via positional encoding;
Plenoxels~\cite{yu2021plenoxels}, DVGO~\cite{SunSC22}, and instantNGP~\cite{muller2022instant}, which are grid-based methods.

\paragraph{Results.}
We report our results in \Figure{NVS} and \Table{NVS}.
Our method provides similar performance as other methods, but with a much smaller model size.

\subsection{Ablation Study}
\label{sec:ablation}

To justify the design choices of our method we explore three variants of our method: 
our method where only Grid features are used as `Only Grid';
our method with Grid and the Fourier features encoding as `Grid+FF';
and finally when only using the MLP architecture for composition without the grid as `Only MLP'.
For a fair evaluation of the effects of the MLP part, we adjust the `Only MLP' model to possess similar number of trainable parameters as the full model.
We report our results for the `Tokyo' image in \Figure{ablation_component}.
As shown, all variants perform significantly worse. 
Interestingly, simply applying Fourier Features to the grid does not help, demonstrating the proposed MLP architecture is also necessary to achieve its potential.

\section{Conclusions}

We have proposed the neural Fourier filter bank, inspired by wavelets, that provide high-quality reconstruction with more compact models.
We have shown that taking into account both the space and frequency is critical when decomposing the original signal as neural field grids.
Our method provides the best trade-off between quality and model compactness for 2D image reconstruction, 3D shape representation, and novel-view synthesis via NeRF.

\section*{Acknowledgement}

This work was supported by the Natural Sciences and Engineering Research Council of Canada (NSERC) Discovery Grant, Digital Research Alliance of Canada, and by Advanced Research Computing at the University of British Columbia.

{\small
\bibliographystyle{ieee_fullname}
\bibliography{macros,egbib}

\begin{thebibliography}{10}\itemsep=-1pt

\bibitem{barron2021mip}
Jonathan~T Barron, Ben Mildenhall, Matthew Tancik, Peter Hedman, Ricardo
  Martin-Brualla, and Pratul~P Srinivasan.
\newblock {Mip-NeRF: A Multiscale Representation for Anti-Aliasing Neural
  Radiance Fields}.
\newblock In {\em Int. Conf. Comput. Vis.}, 2021.

\bibitem{bello2019attention}
Irwan Bello, Barret Zoph, Ashish Vaswani, Jonathon Shlens, and Quoc~V Le.
\newblock {Attention Augmented Convolutional Networks}.
\newblock In {\em Int. Conf. Comput. Vis.}, 2019.

\bibitem{benbarka2022seeing}
Nuri Benbarka, Timon H{\"o}fer, Andreas Zell, et~al.
\newblock Seeing implicit neural representations as fourier series.
\newblock 2022.

\bibitem{chen2021learning}
Yinbo Chen, Sifei Liu, and Xiaolong Wang.
\newblock {Learning Continuous Image Representation with Local Implicit Image
  Function}.
\newblock In {\em IEEE Conf. Comput. Vis. Pattern Recog.}, 2021.

\bibitem{chen2022mobilenerf}
Zhiqin Chen, Thomas Funkhouser, Peter Hedman, and Andrea Tagliasacchi.
\newblock {MobileNeRF: Exploiting the Polygon Rasterization Pipeline for
  Efficient Neural Field Rendering on Mobile Architectures}.
\newblock {\em ArXiv preprint}, 2022.

\bibitem{chen2019learning}
Zhiqin Chen and Hao Zhang.
\newblock {Learning Implicit Fields for Generative Shape Modeling}.
\newblock In {\em IEEE Conf. Comput. Vis. Pattern Recog.}, 2019.

\bibitem{chen2022tensorf}
{Chen, Anpei and Xu, Zexiang and Geiger, Andreas and Yu, Jingyi and Su, Hao}.
\newblock {TensoRF: Tensorial Radiance Fields}.
\newblock In {\em Eur. Conf. Comput. Vis.}, 2022.

\bibitem{de2016compression}
Ricardo~L De~Queiroz and Philip~A Chou.
\newblock {Compression of 3D point clouds using a region-adaptive hierarchical
  transform}.
\newblock {\em IEEE Trans. Image Process.}, 2016.

\bibitem{deng2020cvxnet}
Boyang Deng, Kyle Genova, Soroosh Yazdani, Sofien Bouaziz, Geoffrey Hinton, and
  Andrea Tagliasacchi.
\newblock {CvxNet: Learnable Convex Decomposition}.
\newblock In {\em IEEE Conf. Comput. Vis. Pattern Recog.}, 2020.

\bibitem{dupont2022coin++}
Emilien Dupont, Hrushikesh Loya, Milad Alizadeh, Adam Goli{\'n}ski, Yee~Whye
  Teh, and Arnaud Doucet.
\newblock {COIN++: Neural Compression Across Modalities}.
\newblock {\em ArXiv preprint}, 2022.

\bibitem{tineuvox}
Jiemin Fang, Taoran Yi, Xinggang Wang, Lingxi Xie, Xiaopeng Zhang, Wenyu Liu,
  Matthias Nie{\ss}ner, and Qi Tian.
\newblock {Fast Dynamic Radiance Fields with Time-Aware Neural Voxels}.
\newblock {\em ACM SIGGRAPH}, 2022.

\bibitem{fathony2021multiplicative}
Rizal Fathony, Anit~Kumar Sahu, Devin Willmott, and J~Zico Kolter.
\newblock Multiplicative filter networks.
\newblock In {\em Int. Conf. Learn. Represent.}, 2021.

\bibitem{gal2021swagan}
Rinon Gal, Dana~Cohen Hochberg, Amit Bermano, and Daniel Cohen-Or.
\newblock {SWAGAN: A Style-based Wavelet-driven Generative Model}.
\newblock {\em ACM Trans. Graph.}, 2021.

\bibitem{gao2016hybrid}
Xing Gao and Hongkai Xiong.
\newblock {A Hybrid Wavelet Convolution Network with Sparse-Coding for Image
  Super-Resolution}.
\newblock In {\em IEEE Int. Conf. Image Process.}, 2016.

\bibitem{gauthier2022parametric}
Shanel Gauthier, Benjamin Th{\'e}rien, Laurent Alsene-Racicot, Muawiz
  Chaudhary, Irina Rish, Eugene Belilovsky, Michael Eickenberg, and Guy Wolf.
\newblock Parametric scattering networks.
\newblock In {\em IEEE Conf. Comput. Vis. Pattern Recog.}, 2022.

\bibitem{genova2020local}
Kyle Genova, Forrester Cole, Avneesh Sud, Aaron Sarna, and Thomas Funkhouser.
\newblock {Local Deep Implicit Functions for 3D Shape}.
\newblock In {\em IEEE Conf. Comput. Vis. Pattern Recog.}, 2020.

\bibitem{genova2019learning}
Kyle Genova, Forrester Cole, Daniel Vlasic, Aaron Sarna, William~T Freeman, and
  Thomas Funkhouser.
\newblock {Learning Shape Templates with Structured Implicit Functions}.
\newblock {\em Int. Conf. Comput. Vis.}, 2019.

\bibitem{hertz2021sape}
Amir Hertz, Or Perel, Raja Giryes, Olga Sorkine-Hornung, and Daniel Cohen-Or.
\newblock {SAPE: Spatially-Adaptive Progressive Encoding for Neural
  Optimization}.
\newblock {\em Adv. Neural Inform. Process. Syst.}, 2021.

\bibitem{hore2010image}
Alain Hore and Djemel Ziou.
\newblock {Image Quality Metrics: PSNR vs. SSIM}.
\newblock In {\em Int. Conf. Pattern Recog.}, 2010.

\bibitem{hu2022efficientnerf}
Tao Hu, Shu Liu, Yilun Chen, Tiancheng Shen, and Jiaya Jia.
\newblock {EfficientNeRF: Efficient Neural Radiance Fields }.
\newblock In {\em IEEE Conf. Comput. Vis. Pattern Recog.}, 2022.

\bibitem{huang2019wavelet}
Huaibo Huang, Ran He, Zhenan Sun, and Tieniu Tan.
\newblock {Wavelet Domain Generative Adversarial Network for Multi-scale Face
  Hallucination}.
\newblock {\em Int. J. Comput. Vis.}, 2019.

\bibitem{isik2021lvac}
Berivan Isik, Philip Chou, Sung~Jin Hwang, Nicholas Johnston, and George
  Toderici.
\newblock {LVAC: Learned volumetric attribute compression for point clouds
  using coordinate based networks}.
\newblock {\em Frontiers in Signal Processing}, 2021.

\bibitem{jiang2020local}
Chiyu Jiang, Avneesh Sud, Ameesh Makadia, Jingwei Huang, Matthias Nie{\ss}ner,
  Thomas Funkhouser, et~al.
\newblock {Local Implicit Grid Representations for 3D Scenes}.
\newblock In {\em IEEE Conf. Comput. Vis. Pattern Recog.}, 2020.

\bibitem{kang2017deep}
Eunhee Kang, Junhong Min, and Jong~Chul Ye.
\newblock {A deep convolutional neural network using directional wavelets for
  low-dose X-ray CT reconstruction}.
\newblock {\em Medical physics}, 2017.

\bibitem{kingma2014adam}
Diederik~P Kingma and Jimmy Ba.
\newblock {Adam: A method for stochastic optimization}.
\newblock {\em Int. Conf. Learn. Represent.}, 2014.

\bibitem{levoy1996light}
Marc Levoy and Pat Hanrahan.
\newblock {Light Field Rendering}.
\newblock In {\em SIGGRAPH}, 1996.

\bibitem{liang2021reproducing}
Senwei Liang, Liyao Lyu, Chunmei Wang, and Haizhao Yang.
\newblock Reproducing activation function for deep learning.
\newblock 2021.

\bibitem{lindell2021bacon}
David~B Lindell, Dave Van~Veen, Jeong~Joon Park, and Gordon Wetzstein.
\newblock {BACON: Band-limited Coordinate Networks for Multiscale Scene
  Representation}.
\newblock {\em IEEE Conf. Comput. Vis. Pattern Recog.}, 2022.

\bibitem{liu2020wavelet}
Lin Liu, Jianzhuang Liu, Shanxin Yuan, Gregory Slabaugh, Ale{\v{s}} Leonardis,
  Wengang Zhou, and Qi Tian.
\newblock {Wavelet-Based Dual-Branch Network for Image Demoireing}.
\newblock In {\em Eur. Conf. Comput. Vis.}, 2020.

\bibitem{liu2019multi}
Pengju Liu, Hongzhi Zhang, Wei Lian, and Wangmeng Zuo.
\newblock {Multi-level Wavelet Convolutional Neural Networks}.
\newblock {\em IEEE Access}, 2019.

\bibitem{liu2018intriguing}
Rosanne Liu, Joel Lehman, Piero Molino, Felipe Petroski~Such, Eric Frank, Alex
  Sergeev, and Jason Yosinski.
\newblock {An Intriguing Failing of Convolutional Neural Networks and the
  CoordConv Solution}.
\newblock {\em Adv. Neural Inform. Process. Syst.}, 2018.

\bibitem{liu2019attribute}
Yunfan Liu, Qi Li, and Zhenan Sun.
\newblock {Attribute-aware Face Aging with Wavelet-based Generative Adversarial
  Networks}.
\newblock In {\em IEEE Conf. Comput. Vis. Pattern Recog.}, 2019.

\bibitem{martel2021acorn}
Julien~N.P. Martel, David~B. Lindell, Connor~Z. Lin, Eric~R. Chan, Marco
  Monteiro, and Gordon Wetzstein.
\newblock {ACORN: Adaptive Coordinate Networks for Neural Scene
  Representation}.
\newblock {\em SIGGRAPH}, 2021.

\bibitem{mehta2021modulated}
Ishit Mehta, Micha{\"e}l Gharbi, Connelly Barnes, Eli Shechtman, Ravi
  Ramamoorthi, and Manmohan Chandraker.
\newblock {Modulated Periodic Activations for Generalizable Local Functional
  Representations}.
\newblock In {\em Int. Conf. Comput. Vis.}, 2021.

\bibitem{mescheder2019occupancy}
Lars Mescheder, Michael Oechsle, Michael Niemeyer, Sebastian Nowozin, and
  Andreas Geiger.
\newblock {Occupancy Networks: Learning 3D Reconstruction in Function Space}.
\newblock In {\em IEEE Conf. Comput. Vis. Pattern Recog.}, 2019.

\bibitem{mildenhall2020nerf}
Ben Mildenhall, Pratul~P. Srinivasan, Matthew Tancik, Jonathan~T. Barron, Ravi
  Ramamoorthi, and Ren Ng.
\newblock {NeRF: Representing Scenes as Neural Radiance Fields for View
  Synthesis}.
\newblock In {\em Eur. Conf. Comput. Vis.}, 2020.

\bibitem{muller2022instant}
Thomas M{\"u}ller, Alex Evans, Christoph Schied, and Alexander Keller.
\newblock {Instant Neural Graphics Primitives with a Multiresolution Hash
  Encoding}.
\newblock {\em SIGGRAPH}, 2022.

\bibitem{park2019deepsdf}
Jeong~Joon Park, Peter Florence, Julian Straub, Richard Newcombe, and Steven
  Lovegrove.
\newblock {DeepSDF: Learning Continuous Signed Distance Functions for Shape
  Representation}.
\newblock In {\em IEEE Conf. Comput. Vis. Pattern Recog.}, 2019.

\bibitem{pytorch}
Adam Paszke, Sam Gross, Francisco Massa, Adam Lerer, James Bradbury, Gregory
  Chanan, Trevor Killeen, Zeming Lin, Natalia Gimelshein, Luca Antiga, Alban
  Desmaison, Andreas Kopf, Edward Yang, Zachary DeVito, Martin Raison, Alykhan
  Tejani, Sasank Chilamkurthy, Benoit Steiner, Lu Fang, Junjie Bai, and Soumith
  Chintala.
\newblock {PyTorch: An Imperative Style, High-Performance Deep Learning
  Library}.
\newblock In {\em Adv. Neural Inform. Process. Syst.}, 2019.

\bibitem{peleg2019net}
Tomer Peleg, Pablo Szekely, Doron Sabo, and Omry Sendik.
\newblock Im-net for high resolution video frame interpolation.
\newblock In {\em IEEE Conf. Comput. Vis. Pattern Recog.}, 2019.

\bibitem{peng2020convolutional}
Songyou Peng, Michael Niemeyer, Lars Mescheder, Marc Pollefeys, and Andreas
  Geiger.
\newblock {Convolutional Occupancy Networks}.
\newblock In {\em Eur. Conf. Comput. Vis.}, 2020.

\bibitem{phung2022wavelet}
Hao Phung, Quan Dao, and Anh Tran.
\newblock Wavelet diffusion models are fast and scalable image generators.
\newblock {\em arXiv preprint}, 2022.

\bibitem{qi2016volumetric}
Charles~R Qi, Hao Su, Matthias Nie{\ss}ner, Angela Dai, Mengyuan Yan, and
  Leonidas~J Guibas.
\newblock {Volumetric and Multi-View CNNs for Object Classification on 3D
  Data}.
\newblock In {\em IEEE Conf. Comput. Vis. Pattern Recog.}, 2016.

\bibitem{rho2023masked}
Daniel Rho, Byeonghyeon Lee, Seungtae Nam, Joo~Chan Lee, Jong~Hwan Ko, and
  Eunbyung Park.
\newblock Masked wavelet representation for compact neural radiance fields.
\newblock 2023.

\bibitem{yu2021plenoxels}
{Sara Fridovich-Keil and Alex Yu}, Matthew Tancik, Qinhong Chen, Benjamin
  Recht, and Angjoo Kanazawa.
\newblock {Plenoxels: Radiance Fields without Neural Networks}.
\newblock In {\em IEEE Conf. Comput. Vis. Pattern Recog.}, 2022.

\bibitem{shannon1949communication}
Claude~E Shannon.
\newblock {Communication in the Presence of Noise}.
\newblock {\em Proceedings of the IRE}, 1949.

\bibitem{sitzmann2020implicit}
Vincent Sitzmann, Julien Martel, Alexander Bergman, David Lindell, and Gordon
  Wetzstein.
\newblock {Implicit Neural Representations with Periodic Activation Functions}.
\newblock {\em Adv. Neural Inform. Process. Syst.}, 2020.

\bibitem{sitzmann2019scene}
Vincent Sitzmann, Michael Zollh{\"o}fer, and Gordon Wetzstein.
\newblock {Scene Representation Networks: Continuous 3D-Structure-Aware Neural
  Scene Representations}.
\newblock {\em Adv. Neural Inform. Process. Syst.}, 2019.

\bibitem{skorokhodov2021adversarial}
Ivan Skorokhodov, Savva Ignatyev, and Mohamed Elhoseiny.
\newblock {Adversarial Generation of Continuous Images}.
\newblock In {\em IEEE Conf. Comput. Vis. Pattern Recog.}, 2021.

\bibitem{SunSC22}
Cheng Sun, Min Sun, and Hwann{-}Tzong Chen.
\newblock {Direct Voxel Grid Optimization: Super-fast Convergence for Radiance
  Fields Reconstruction}.
\newblock In {\em IEEE Conf. Comput. Vis. Pattern Recog.}, 2022.

\bibitem{takikawa2021neural}
Towaki Takikawa, Joey Litalien, Kangxue Yin, Karsten Kreis, Charles Loop, Derek
  Nowrouzezahrai, Alec Jacobson, Morgan McGuire, and Sanja Fidler.
\newblock {Neural Geometric Level of Detail: Real-time Rendering with Implicit
  3D Shapes}.
\newblock In {\em IEEE Conf. Comput. Vis. Pattern Recog.}, 2021.

\bibitem{tancik2022block}
Matthew Tancik, Vincent Casser, Xinchen Yan, Sabeek Pradhan, Ben Mildenhall,
  Pratul~P Srinivasan, Jonathan~T Barron, and Henrik Kretzschmar.
\newblock {Block-NeRF: Scalable Large Scene Neural View Synthesis}.
\newblock In {\em IEEE Conf. Comput. Vis. Pattern Recog.}, 2022.

\bibitem{tancik2020fourier}
Matthew Tancik, Pratul Srinivasan, Ben Mildenhall, Sara Fridovich-Keil, Nithin
  Raghavan, Utkarsh Singhal, Ravi Ramamoorthi, Jonathan Barron, and Ren Ng.
\newblock {Fourier Features Let Networks Learn High Frequency Functions in Low
  Dimensional Domains}.
\newblock {\em Adv. Neural Inform. Process. Syst.}, 2020.

\bibitem{teschner2003optimized}
Matthias Teschner, Bruno Heidelberger, Matthias M{\"u}ller, Danat Pomerantes,
  and Markus~H Gross.
\newblock {Optimized Spatial Hashing for Collision Detection of Deformable
  Objects}.
\newblock In {\em VMV}, 2003.

\bibitem{vaswani2017attention}
Ashish Vaswani, Noam Shazeer, Niki Parmar, Jakob Uszkoreit, Llion Jones,
  Aidan~N Gomez, {\L}ukasz Kaiser, and Illia Polosukhin.
\newblock {Attention Is All You Need}.
\newblock {\em Adv. Neural Inform. Process. Syst.}, 2017.

\bibitem{wang2020multi}
Jianyi Wang, Xin Deng, Mai Xu, Congyong Chen, and Yuhang Song.
\newblock {Multi-level Wavelet-based Generative Adversarial Network for
  Perceptual Quality Enhancement of Compressed Video}.
\newblock In {\em Eur. Conf. Comput. Vis.}, 2020.

\bibitem{williams2018wavelet}
Travis Williams and Robert Li.
\newblock {Wavelet Pooling for Convolutional Neural Networks}.
\newblock In {\em Int. Conf. Learn. Represent.}, 2018.

\bibitem{xiangli2021citynerf}
Yuanbo Xiangli, Linning Xu, Xingang Pan, Nanxuan Zhao, Anyi Rao, Christian
  Theobalt, Bo Dai, and Dahua Lin.
\newblock {CityNeRF: Building NeRF at City Scale}.
\newblock {\em Eur. Conf. Comput. Vis.}, 2022.

\bibitem{nfsurvey}
Yiheng Xie, Towaki Takikawa, Shunsuke Saito, Or Litany, Shiqin Yan, Numair
  Khan, Federico Tombari, James Tompkin, Vincent Sitzmann, and Srinath Sridhar.
\newblock {Neural Fields in Visual Computing and Beyond}.
\newblock {\em Computer Graphics Forum}, 2022.

\bibitem{yin20213dstylenet}
Kangxue Yin, Jun Gao, Maria Shugrina, Sameh Khamis, and Sanja Fidler.
\newblock {3DStyleNet: Creating 3D Shapes with Geometric and Texture Style
  Variations}.
\newblock In {\em Int. Conf. Comput. Vis.}, 2021.

\bibitem{yoo2019photorealistic}
Jaejun Yoo, Youngjung Uh, Sanghyuk Chun, Byeongkyu Kang, and Jung-Woo Ha.
\newblock {Photorealistic Style Transfer via Wavelet Transforms}.
\newblock In {\em Int. Conf. Comput. Vis.}, 2019.

\bibitem{yuce2022structured}
Gizem Y{\"u}ce, Guillermo Ortiz-Jim{\'e}nez, Beril Besbinar, and Pascal
  Frossard.
\newblock {A Structured Dictionary Perspective on Implicit Neural
  Representations}.
\newblock In {\em IEEE Conf. Comput. Vis. Pattern Recog.}, 2022.

\bibitem{zhang2018unreasonable}
Richard Zhang, Phillip Isola, Alexei~A Efros, Eli Shechtman, and Oliver Wang.
\newblock {The Unreasonable Effectiveness of Deep Features as a Perceptual
  Metric}.
\newblock In {\em IEEE Conf. Comput. Vis. Pattern Recog.}, 2018.

\bibitem{zhang2022implicit}
Yunfan Zhang, Ties van Rozendaal, Johann Brehmer, Markus Nagel, and Taco Cohen.
\newblock {Implicit Neural Video Compression}.
\newblock In {\em Int. Conf. Learn. Represent. Workshop}, 2022.

\end{thebibliography}
}

\clearpage
\appendix
\renewcommand\thefigure{\Alph{figure}}
\setcounter{figure}{0}
\renewcommand\thetable{\Alph{table}}
\setcounter{table}{0}
\setcounter{footnote}{0}
\twocolumn[
\centering
\Large
\textbf{Neural Fourier Filter Bank} \\
\vspace{0.5em} (Supplementary Material) \\
\vspace{1.0em}
]

\section{Experiment details}
\label{sec:supp_results}

\subsection{More details about Fourier grid features} 
\label{sec:supp_grid_feats}

In Sec.~3.1 of the main text, the multi-level Fourier grid features are defined to compute the continuous mapping between the input coordinate $\bx \in \real^n$ and the $m$ dimension feature space.
Following instantNGP, we set the base resolution $\gridres_{min}$ and a scaling coefficient $c_{g}$ between adjacent levels to define the resolution for a certain level $l$ as:
\begin{equation}
\gridres_l = \gridres_{min} \cdot c_{g}^l
,
\end{equation}
where the level index $l$ starts from 0.
We adjust the total number of levels to balance the 
ability to model fine details and the complexity of the model itself.
To compute the variance values that we use to initialize
the Fourier features, we apply a similar scaling strategy:
\begin{equation}
\sigma_{l} = \sigma_{min} \cdot c_{f}^l
,
\end{equation}
where $\sigma_{min}$ and $c_{f}$ represent the base variance value and its corresponding scaling coefficient.
Roughly, we set $\sigma_{min}= \sqrt[2]{N_{min}}$ and $c_g \approx c_f$.
However, the optimal choice of these values is circumstantial and we modify 
$\gridres_{min}$, $c_{g}$, $\sigma_{min}$ and $c_{f}$
for each task.

\subsection{2D image fitting} 
\label{sec:supp_img_fitting}

To roughly match the model capacity used by other methods, for the `Tokyo' image we use fully-connected layers with 96 neurons, and for the `Einstein' image 256.
All fully connected layers are using sine activations as previously described in the main text.
Additionally, for the `Tokyo' image we use $\gridres_{min}=64$, $c_{g}=1.5$, $\sigma_{min}=5.0$ and $c_{f}=2.0$, and for the `Einstein' image we use $\gridres_{min}=64$, $c_{g}=2.0$, $\sigma_{min}=10.0$ and $c_{f}=2.0$. 

For both images, we train our network with 50,000 iterations to ensure full convergence---our method already converges after 20,000 iterations.
To well-reconstruct complex high-frequency signals, we set $\alpha_i$ in Eq.~(5) to 100.

\subsection{3D SDF regression} 
\label{sec:supp_shape_regression}

For this task, we train our network for 50,000 iterations on 26 million sampled points with a batch size of 49,152, to maximize GPU memory utilization.
As the SDF has varying level-of-detail---\eg, smooth regions can be very smooth, while detailed regions can have high-frequency detail---we set the number of levels to five for the Fourier grid feature.
For each level, we use fully-connected layers with 256 neurons.
We further set 
$\gridres_{min}=8$, $c_{g}=1.3$, $\sigma_{min}=5$ and $c_{f}=1.2$.
For both shapes in Tab.~2, we choose $\alpha_i=45$, which we empirically found to provide the best balance between high- and low-frequency details for this task.

\subsection{Neural radiance field} 
\label{sec:supp_nerf}

For this task, we closely follow the experimental setup of InstantNGP, including the four levels for the grid.
For our method, we use 128 neurons to match a similar model capacity as the baseline.
We further set $\alpha_i=20.0$, $\gridres_{min}=64$, $c_{g}=2.0$, $\sigma_{min}=8.0, c_{f}=1.4$ to balance model complexity and synthesis quality.

\subsection{Preparing SDF data for SDF regression} 
\label{sec:supp_sdf_computation}

To obtain the ground-truth SDF values, we use pysdf\footnote{Github link: \url{https://github.com/sxyu/sdf}}.
We use the original mesh files and normalize them into a unit sphere to standardize shapes.
When training each model, for each batch,
we sample 49152 points for training where $20\%$ of the points are sampled uniformly within the volume, $30\%$ of the points are sampled near the shape surface, and the rest are sampled directly on the surface.

\begin{table}[t]

\begin{center}

\resizebox{\linewidth}{!}{
\setlength{\tabcolsep}{12pt}
\begin{tabular}{@{}l c c c c c c c c c c@{}}
    \toprule
     & \multicolumn{3}{c}{$T=2^{17}$} & \multicolumn{3}{c}{$T=2^{19}$} & \multicolumn{3}{c}{$T=2^{21}$}\\
    \cmidrule(l){2-4} \cmidrule(l){5-7} \cmidrule(l){8-10} 
    & $L=8$ & $L=10$ & $L=12$ & $L=8$ & $L=10$ & $L=12$ & $L=8$ & $L=10$ & $L=12$ \\
    \midrule
    InstantNGP  & 28.18 & 30.89 & 31.93 & 29.38 & 33.37 & 36.41 & 30.41 & 36.37 & 41.28\\
    Ours & 30.31 & 32.86 & 33.82 & 31.28 & 34.53 & 37.56 & 31.43 & 36.86 & 41.36 \\
    \bottomrule
\end{tabular}
}
\end{center}
\vspace{-0.5em}
\caption{
\textbf{Performance under varying $T$ and $L$} --
Our method shows higher PSNR values for `Tokyo' image with various $T$ and $L$ settings.
\label{tab:supp_hyperparameter}
\vspace{-0.5em}
}
\end{table}

\subsection{Experimental setting for InstantNGP} 
\label{sec:supp_ngp_settings}

Generally, our choices are based \cite[Sec.~3.]{muller2022instant}.
As shown in \cite[Fig.~5.]{muller2022instant}, $F{=}2$ and $L{=}16$ are good choices for the feature dimension $F$ and feature level $L$.
For the hash table size $T$, we choose $2^{19}$ as it is when the performance starts being throttled as shown in \cite[Fig.~4.]{muller2022instant}.
For the `Einstein' image in Tab.~1 of the main text, we reduced the model's capacity as the image is simpler.

In addition, we use various settings for $T$ and $L$ for
both InstantNGP and our method and report the results for `Tokyo' image in Tab.~\ref{tab:supp_hyperparameter}.
Regardless of the hyperparameter settings, our method outperforms InstantNGP consistently.

\section{More ablation studies}
\label{sec:supp_ab_test}

As discussed in Sec.~3 of the main text,
our key idea is the Fourier grid features, and the wavelet-inspired composition.
Here, we further justify our design choices based on experiments.

\paragraph{The effect of grid resolutions.}
In Sec.~1., we discuss how grid resolution relates to what frequency range a model can reconstruct. 
In \Figure{ablation_grid_size_base}, we illustrate that this is indeed the case by varying $\gridres_{min}$.
Also in \Figure{ablation_grid_size_scale}, we show how the scaling factor $c_g$ affects final results.
As expected, whether fine details are preserved or not depends highly on the two parameters.

\paragraph{The effect of the Fourier feature variance.}
\begin{figure}
    \centering
    \includegraphics[width=\linewidth,trim={0cm 14.2cm 20cm 3.0cm},clip]{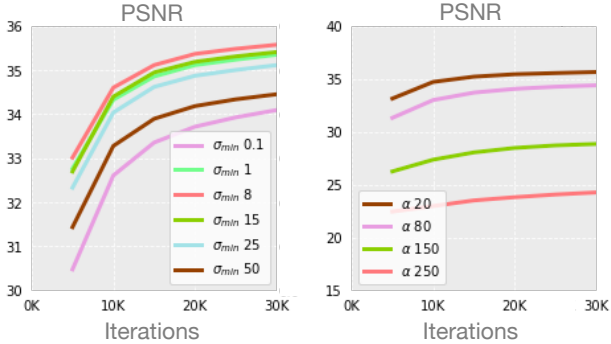}
    \caption{
    Ablation studies for the Fourier feature variance (left) and the scaling factor $\alpha$ in fully-connected layers (right).
    Both parameters highly affect how frequency is dealt with within our framework, and thus require optimal values to be set. 
    These parameters are mostly task dependant.
    }
    \label{fig:ablation_nvs}
\end{figure}
In Sec.~3.1, we discuss how our initialization strategy leads to the natural biasing of frequency components.
Thus, this variance has a strong impact on the performance of the method---too high variance would lead to the method focusing only on high-frequencies, while too low variance would cause the opposite.
Thus, this variance should be selected with care.
In \Figure{ablation_nvs}, we show how the variance $\sigma_{min}$ affects the final reconstruction performance---$\sigma_{min}$ should roughly be in a proper range, as demonstrated by the $\sigma_{min}=1$ and $\sigma_{min}=8$ results.

\paragraph{The effect of the scaling factor $\alpha_i$.}
Similarly, $\alpha_i$ is another parameter that highly impacts how each layer combines grid features and the features from the previous layer.
We set a single global value for all layers for simplicity, and experiment with multiple values to demonstrate its effect in \Figure{ablation_nvs}.
As expected, a properly tuned value is necessary for optimal performance. 
We found this parameter to be highly task dependant.

\paragraph{The effect of Fourier encodings}
We also demonstrate the influences of applying Fourier encodings to the low dimensional grid features by only preserving the Grid+MLP components.
We train this ablated model for the `Tokyo' image which gives the PSNR of 30.48, whereas the full model yields 31.57. 
To further evaluate the effects of activations for grid features, we implement by replacing the sine activation functions with Relu and produce 30.39, highlighting the necessity of current design choices.

\paragraph{The effect of fully-connected layer size.}
The size of the MLP also plays an important role, as it allows for more complex composition of signals coming from different frequencies.
In \Figure{ablation_mlp_size}, we illustrate the importance of the MLP size--- the larger the better, but with an increase in computation and model complexity.

\paragraph{The effect of the Fourier grid level.}
Finally, we demonstrate how the number of Fourier grid levels affects our results.
As expected, we observe in \Figure{ablation_mlp_layer} that the models with higher grid levels consistently provide better results.

\section{More visualization results}
\label{sec:supp_visualization}

In \Figure{supp_albert} and \Figure{supp_tokyo}, we provide more detailed look into the 2D reconstruction results.
Both results provide highly impressive reconstructions, without any discernable differences to the ground truth.

In \Figure{supp_3d}, we further provide the qualitative results for regressing the `Asian Dragon' shape SDF.
Our method and InstantNGP both provide results with very fine details, but ours is more compact.

We provide more qualitative results for novel view synthesis in \Figure{supp_nvs}.
As shown, our method is able to provide synthesis results with both low-frequency details as shown by the `Lego' scene, and high-frequency details as shown by the `Ficus' scene with thin structures.

\begin{table}[t]

\begin{center}

\resizebox{\linewidth}{!}{
\setlength{\tabcolsep}{12pt}
\begin{tabular}{@{}l c c c c c c c c@{}}
    \toprule
     & \multicolumn{4}{c}{2D Fitting} & \multicolumn{3}{c}{3D Fitting}\\
    \cmidrule(l){2-5} \cmidrule(l){6-8} 
    & Size~(MB)$\downarrow$ & PSNR$\uparrow$ & SSIM$\uparrow$ & LPIPS$\downarrow$ & Size~(MB)$\downarrow$ & F-score$\uparrow$    & CD$\downarrow$ \\
    \midrule
    InstantNGP [33]  & 36.0 & 37.93 & 0.9578 & 0.092 & 46.5 & 0.845 & 0.00295\\
    SIREN [41] & 5.2 & 33.35 & 0.9227 & 0.253 & 2.0 & 0.806 & 0.00370 \\
    ModSine [30] & 3.5 & 28.63 & 0.8316 & 0.409 & 12.0 & 0.604 & 0.00386 \\
    \midrule
    Ours$^*$ & 4.1 & 34.64 & 0.9326 & 0.136 & - & - & - \\
    Ours & 10.0 & 38.64 & 0.9672 & 0.064 & 1.4 & 0.833 & 0.00297 \\
    \bottomrule
\end{tabular}
}
\end{center}
\vspace{-1.0em}
\caption{
\textbf{More comparisons} --
With more compact networks, our method can produce competitive or better results compared to baselines.
Our smaller model (Ours$^*$) is achieved by using smaller grid sizes.
\label{tab:supp_comparisons}
\vspace{-1.5em}
}
\end{table}

\section{More comparison results}
\label{sec:supp_comparison}

While hyperparameters differ for each task, we found them to be generally applicable to other scenes for the same task.
In Tab.~\ref{tab:supp_comparisons} with the hyperparameters used in the main text, we compare our method on six high-resolution images with each image having more than 10 million pixels, and ten 3D scenes with complicated geometric details. 
It is clear that the proposed method can consistently achieve better or comparable results with much smaller model size.

\section{Note on comparison with ModSine}
\label{sec:supp_modsine_comparison}

We use the local representation with a tile size of 64 for 2D \&\& 3D signal fitting, under the auto-decoding setup.
For ModSine~\cite{mehta2021modulated}, 
we have taken the network from the official implementation\footnote{Code from \url{https://ishit.github.io/modsine/}} and included it in our training and evaluation code, to keep all training aspects identical to ours.
We note, however, that our results might not have optimal hyperparameter settings, as some of the experimental setups (layer number, layer size, batch size, and learning rate) were chosen by us as they were unavailable in \cite{mehta2021modulated}.

\section{Discussions about runtime}
\label{sec:supp_running_time}

As shown in Tab.~3 of the main text, our current implementation is not utilizing CUDA libraries (e.g. tiny-cuda-nn\footnote{Code from: \url{https://github.com/NVlabs/tiny-cuda-nn}}) in places other than the hash grid, thus slower than InstantNGP as of now.
Our current implementation requires around 13 minutes to train a Blender scene for the NeRF task, whereas InstantNGP takes around 3–4 minutes. 
However, we suspect that with a more efficient implementation, for example with a full CUDA-integrated implementation such as InstantNGP, would greatly accelerate our method, as our method only introduces a few small linear layers and Fourier Feature embedding layers, which should not increase the computation load significantly.
Finally, recall that as shown in Fig.~1 of the main paper, our method converges faster in terms of number of optimization steps than other methods, including InstantNGP.

\section{Limitations and future work}
\label{sec:supp_limitation}

One limitation of our work is that we assume a stationary neural field, which is not conditioned, similar to how InstantNGP is limited.
Thus, a potentially fruitful research direction would be to incorporate recent conditional neural field methods into our framework.
We also notice that all grid-based methods do have issues when modeling very large scenes. 
This is also another potentially interesting research direction.

\section{Broader impact}
\label{sec:supp_impact}

Our work is of fundamental nature and is not immediately linked to any particular application.
However, our method would facilitate efficient neural field representations, which can widen the potential application area of neural fields.
In addition, our method, being more efficient, would reduce the amount of computing and power consumption required for the application of these methods.

\begin{figure*}
    \centering
    \includegraphics[width=\linewidth,trim={0cm 14.2cm 1.4cm 3cm},clip]{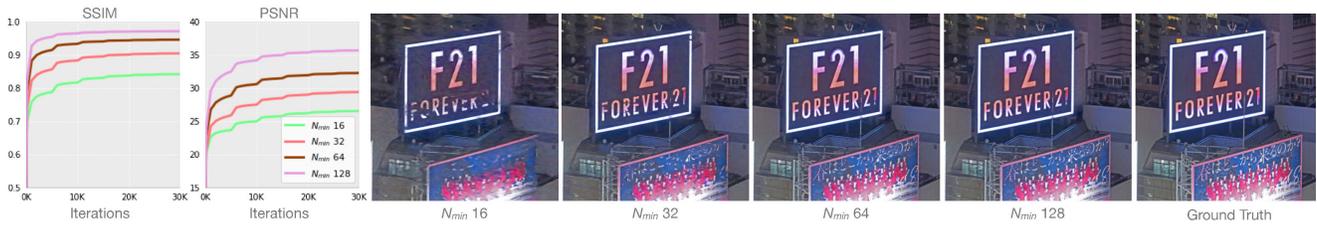}
    \caption{
    The ablation study for the base resolution $N_{min}$. 
    With larger $N_{min}$, 
    fine details are better preserved.
    }
    \label{fig:ablation_grid_size_base}
\end{figure*}
\begin{figure*}
    \centering
    \includegraphics[width=\linewidth,trim={0cm 14.2cm 5.2cm 3cm},clip]{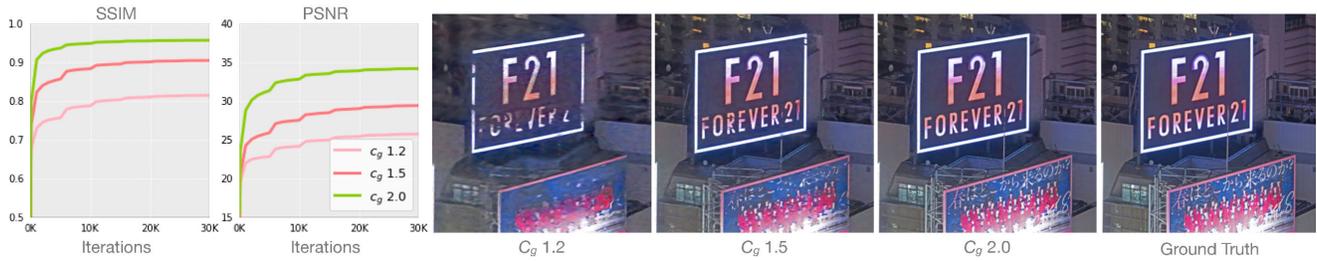}
    \caption{
    The ablation study for the scaling coefficient $c_{g}$. 
    With larger $c_{g}$, 
    reconstruction quality improves, with more fine details being preserved and with higher spatial resolution.
    }
    \label{fig:ablation_grid_size_scale}
\end{figure*}

\begin{figure*}
    \centering
    \includegraphics[width=\linewidth,trim={0cm 14.2cm 1.1cm 2.5cm},clip]{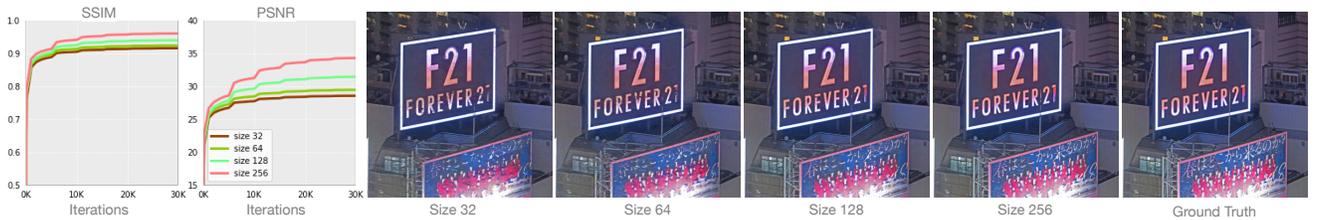}
    \caption{
    The ablation study results for the MLP size. 
    As the MLP size increases, the network becomes better at composing signals from various levels, thus various frequencies, leading to a better final outcome.
    }
    \label{fig:ablation_mlp_size}
\end{figure*}

\begin{figure*}
    \centering
    \includegraphics[width=\linewidth,trim={0cm 14.2cm 1.1cm 2.5cm},clip]{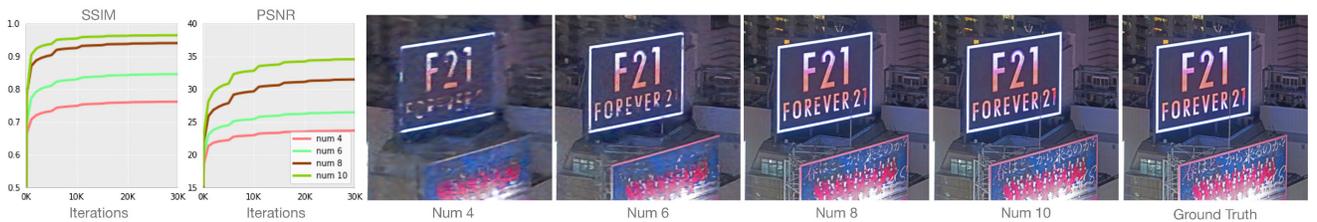}
    \caption{
    The ablation study for the number of levels for the Fourier grid feature.
    More levels lead to a drastic increase in the quality of fine details.
    }
    \label{fig:ablation_mlp_layer}
\end{figure*}

\begin{figure*}
    \centering
    \includegraphics[width=\linewidth,trim={0cm 11.2cm 1cm 2.5cm},clip]{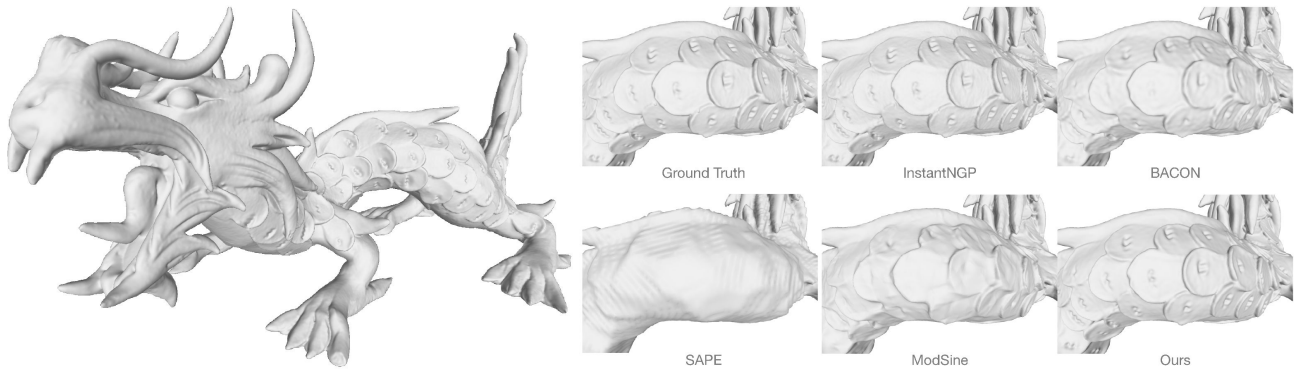}
    \caption{
    3D fitting result for the `Asian Dragon'.
    The left sub-image is the ground truth shape while six zoomed insets are shown on the right for better detail visualizations.
    }
    \label{fig:supp_3d}
\end{figure*}
\begin{figure*}
    \centering
    \includegraphics[width=\linewidth,trim={0cm 10cm 2.9cm 2.9cm},clip]{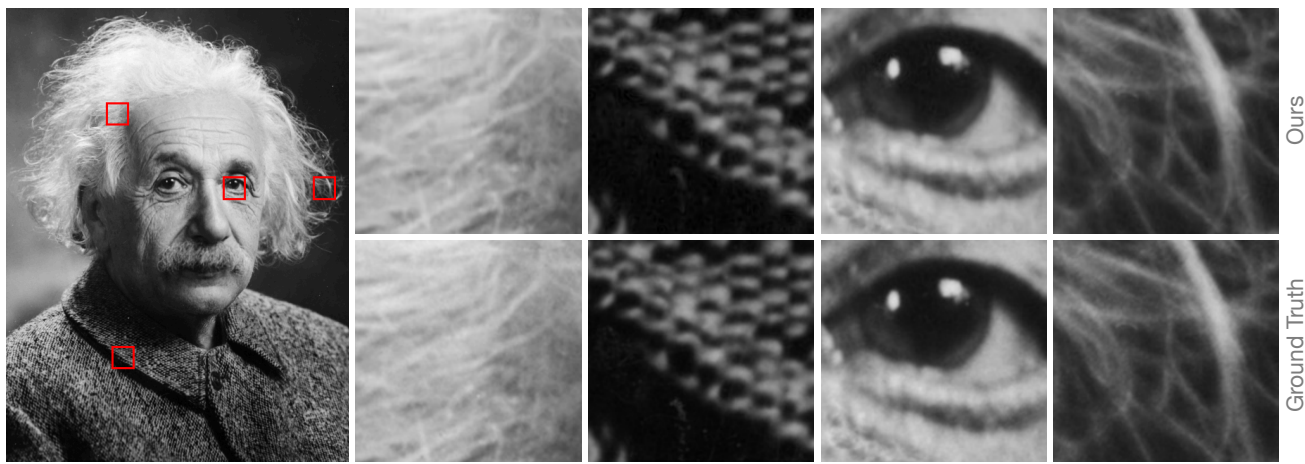}
    \caption{
    2D fitting result for `Einstein' image. 
    Our entire reconstructed image is presented on the left while four close-up views are presented on the right.
    Note how our reconstructions are near-perfect for both coarse and fine details.
    }
    \label{fig:supp_albert}
\end{figure*}
\begin{figure*}
    \centering
    \includegraphics[width=\linewidth,trim={0cm 2.9cm 10cm 2.8cm},clip]{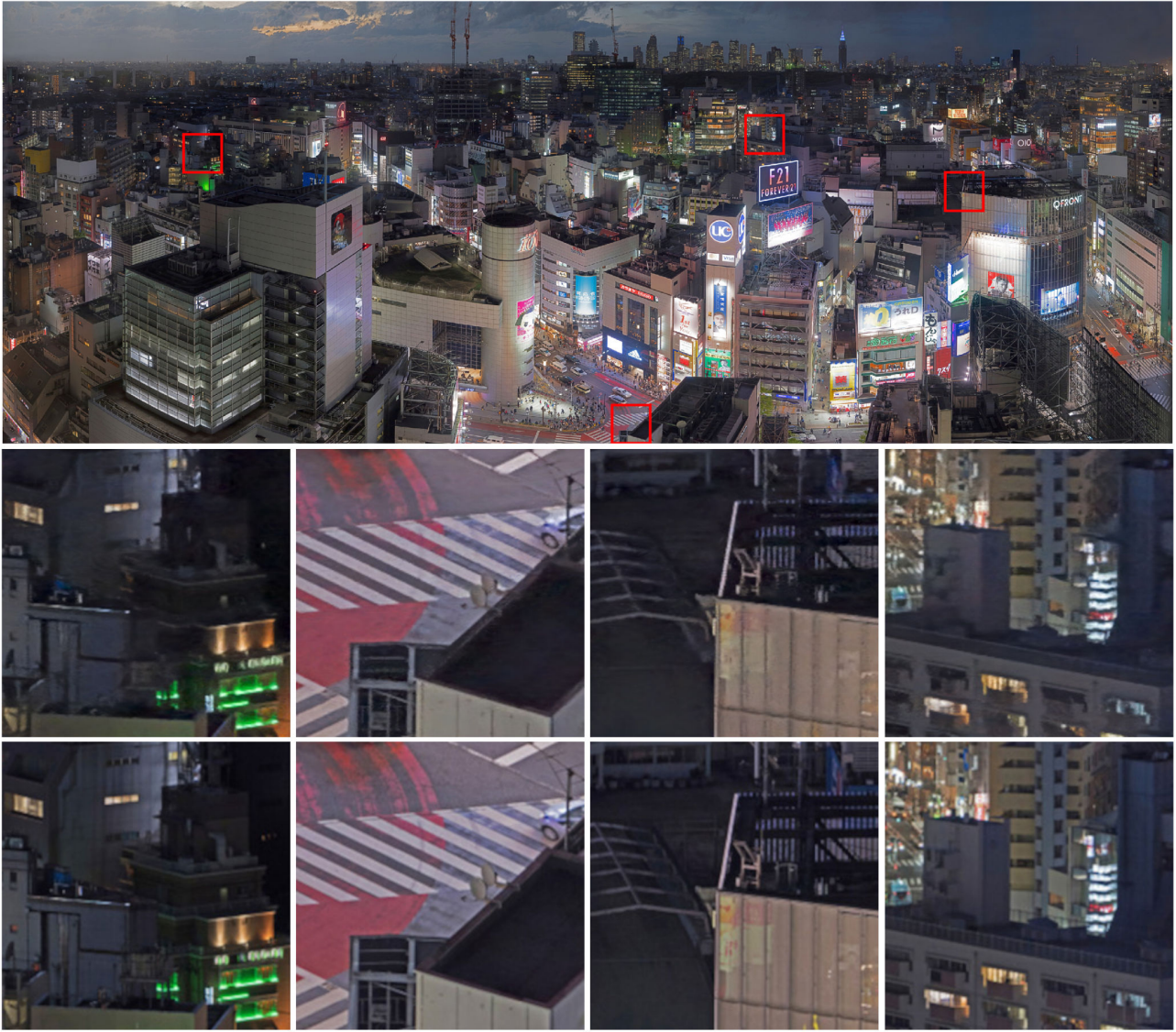}
    \caption{
    2D fitting result for `Tokyo' image. 
    Our entire reconstructed image is presented on the top while four close-up views are presented on the bottom.
    Our method provides near-perfect reconstruction.
    }
    \label{fig:supp_tokyo}
\end{figure*}

\begin{figure*}
    \centering
    \includegraphics[width=0.88\linewidth,trim={0cm 3.7cm 16.5cm 2.8cm},clip]{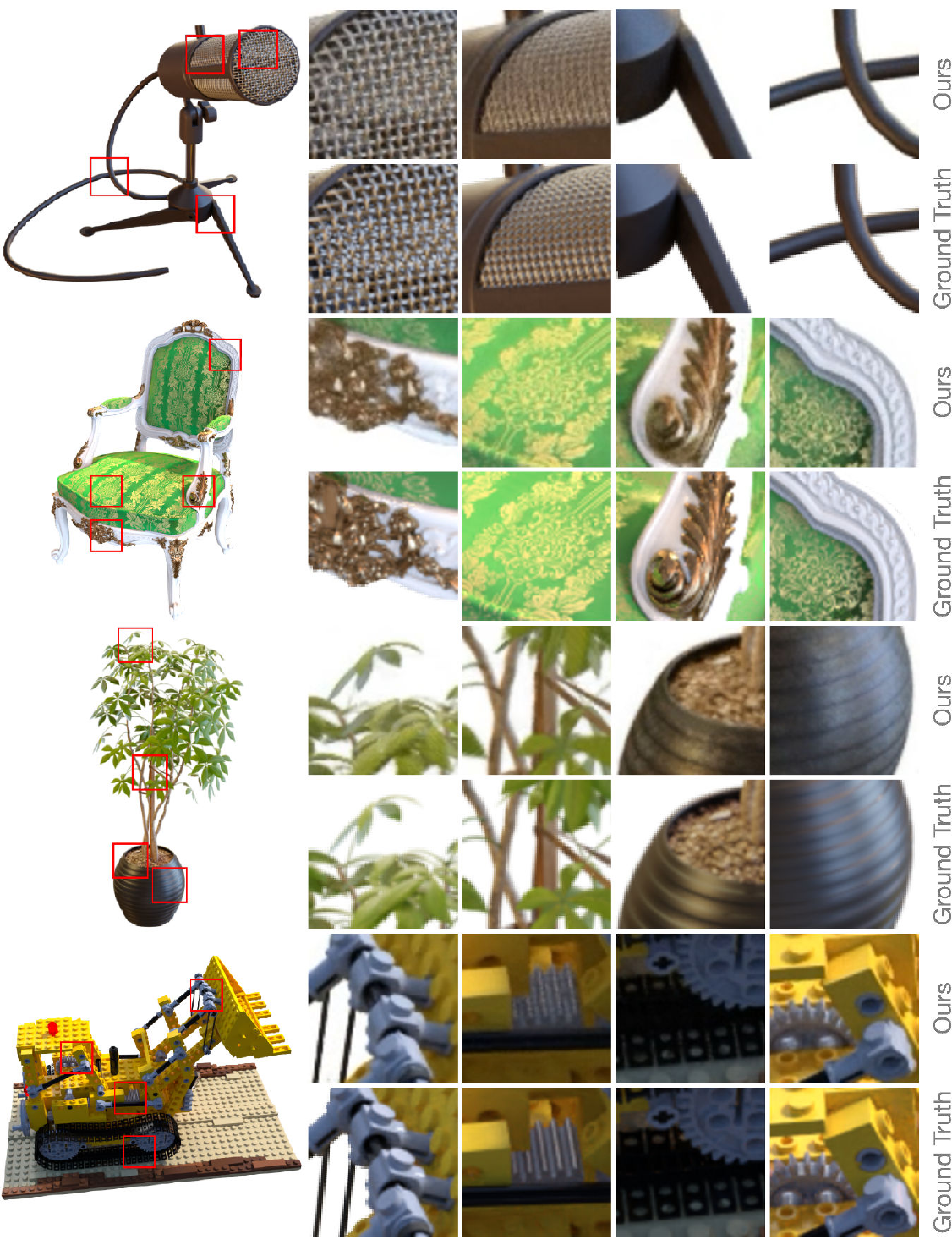}
    \caption{
    Qualitative results for novel view synthesis with neural radiance fields.
    Our method is able to clearly reconstruct the textures (\eg, the chair on $2^{nd}$ row) and the geometric details (\eg the lego on the last row).
    }
    \label{fig:supp_nvs}
\end{figure*}

\end{document}